\documentclass[journal]{IEEEtran}
\ifCLASSINFOpdf
\else
\fi
\usepackage[utf8]{inputenc}
\usepackage{graphicx, subfig}
\usepackage{epstopdf}
\usepackage{amsmath,bm}
\usepackage[numbers,sort&compress]{natbib}
\usepackage{makecell}
\usepackage{multirow}
\usepackage{color}
\usepackage{multicol}
\usepackage{booktabs}
 \usepackage{amssymb}
 \usepackage{setspace}
 \usepackage{mathrsfs}

\begin{document}
%
\title{ A Novel Graph-based Trajectory Predictor with Pseudo Oracle}


\author{Biao~Yang, \textit{Member}, \textit{IEEE}, Guocheng~Yan,  Pin~Wang, \textit{Member}, \textit{IEEE}, Ching-yao~Chan, \textit{Member}, \textit{IEEE}, Xiang~Song, Yang~Chen, \textit{Member},  \textit{IEEE}
\thanks{B. Yang, G. Yan and Y. Chen  are with the Department of Information Science and Engineering, Changzhou University, Changzhou, 213000 China.}
\thanks{P. Wang and C. Chan are with the California PATH, University of California, Berkeley, Richmond, CA, 94804, USA.}
\thanks{X. Song is with the School of Electronic Engineering, Nanjing Xiaozhuang University, Nanjing 211171, China.}
\thanks{Corresponding author: X. Song (songxiang@njxzc.edu.cn)}}

%



\IEEEtitleabstractindextext{%
\begin{abstract}
Pedestrian trajectory prediction in dynamic scenes remains a challenging and critical problem in numerous applications, such as self-driving cars and socially aware robots. Challenges concentrate on capturing pedestrians' motion patterns and social interactions, as well as handling the future uncertainties. Recent studies focus on modeling pedestrians' motion patterns with recurrent neural networks, capturing social interactions with pooling-based or graph-based methods, and handling future uncertainties by using random Gaussian noise as the latent variable. However, they do not integrate specific obstacle avoidance experience (OAE) that may improve prediction performance. For example, pedestrians' future trajectories are always influenced by others in front. Here we propose GTPPO (\textbf{\emph{G}}raph-based \textbf{\emph{T}}rajectory \textbf{\emph{P}}redictor with \textbf{\emph{P}}seudo \textbf{\emph{O}}racle), an encoder-decoder-based method conditioned on pedestrians' future behaviors. Pedestrians' motion patterns are encoded with a long short-term memory unit, which introduces the temporal attention to highlight specific time steps. Their interactions are captured by a graph-based attention mechanism, which draws OAE into the data-driven learning process of graph attention. Future uncertainties are handled by generating multi-modal outputs with an informative latent variable. Such a variable is generated by a novel pseudo oracle predictor, which minimizes the knowledge gap between historical and ground-truth trajectories. Finally, the GTPPO is evaluated on ETH, UCY and Stanford Drone datasets, and the results demonstrate state-of-the-art performance. Besides, the qualitative evaluations show successful cases of handling sudden motion changes in the future. Such findings indicate that GTPPO can peek into the future.
\end{abstract}

\begin{IEEEkeywords}
trajectory prediction, latent variable predictor, social attention, graph attention network, encoder-decoder.
\end{IEEEkeywords}}

\maketitle

\IEEEdisplaynontitleabstractindextext

%
\IEEEpeerreviewmaketitle

\section{Introduction}
Pedestrian trajectory prediction in dynamic scenes remains a critical problem with numerous applications, such as self-driving cars \cite{hong2019rules} and socially aware robots \cite{luber2010people}. For example, a self-driving car can plan a safer path to avoid pedestrian-vehicle collision if future trajectories of pedestrians around can be well predicted. As demonstrated in Fig. 1, pedestrian's future trajectories marked with arrows should be predicted given the historical trajectories marked with lines. Such prediction is challenging for various reasons, e.g., diverse motion patterns and crowd social interactions, including complex human-human and human-objects interactions. Moreover, future uncertainties are difficult to handle due to the multi-modal property of trajectory prediction.

\begin{figure}[!h]
\centering
\includegraphics[width=0.5\textwidth]{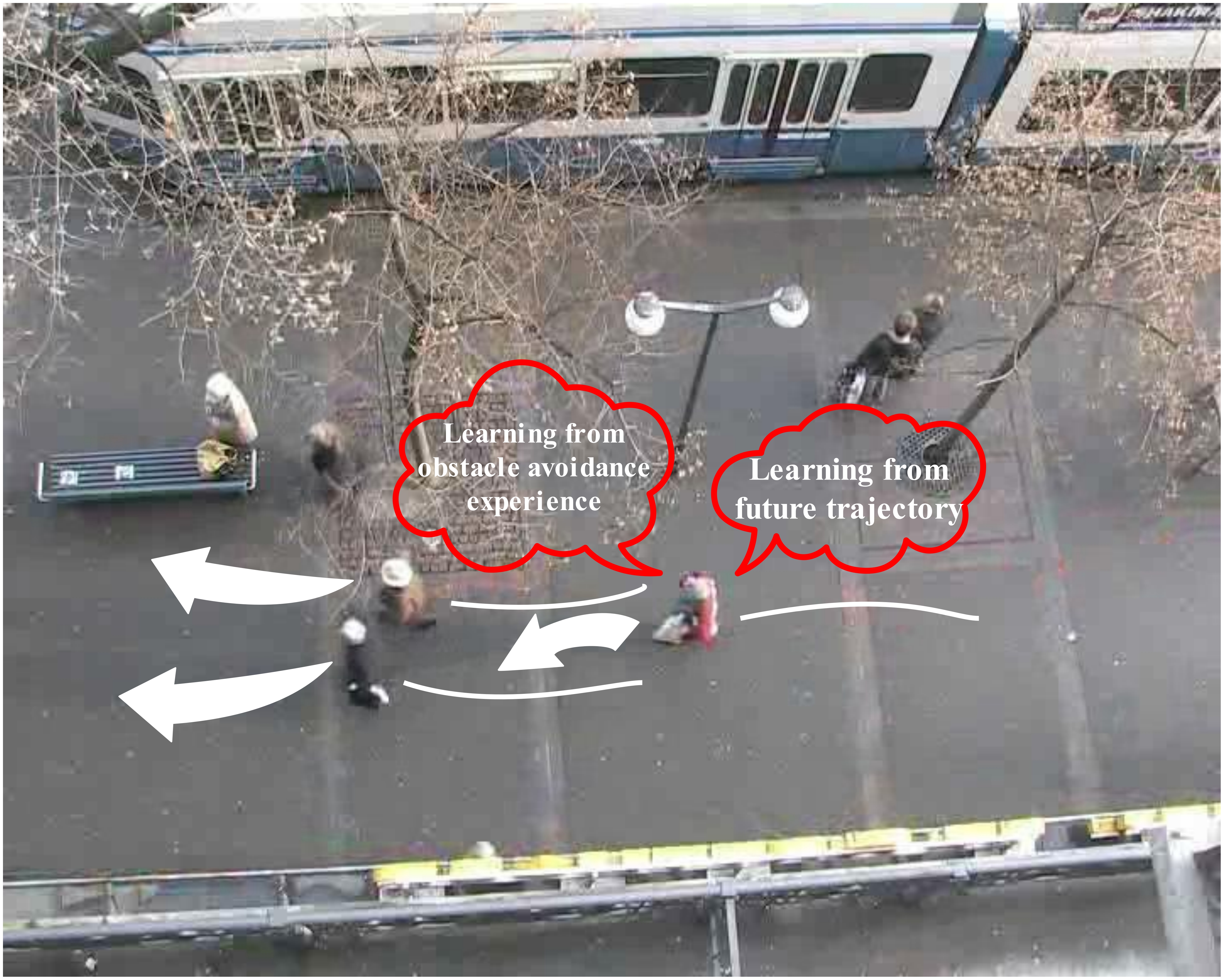}
\caption{Illustration of trajectory prediction. We model a trajectory generator to predict pedestrians' future trajectories based on their historical trajectories. The main characteristics of our method are to learn from the future trajectory and obstacle avoidance experience.}
\label{fig:1}
\end{figure}

Early works related to trajectory prediction always depend on linear, constant velocity and constant acceleration models. However, these simple models show poor performance in complex environments. Later, several stochastic models, including Gaussian mixture regression \cite{li2018generic}, hidden Markov models \cite{wang2018learning}, and dynamic Bayesian networks (DBNs) \cite{kasper2012object} have been introduced to model pedestrians' motion patterns. However, it is challenging to design hand-crafted features due to the diversity of motion patterns. Recent works resort to deep neural networks, e.g., the recurrent neural network (RNN) \cite{lee2017desire}, to automatically encode pedestrians' motion patterns. The Transformer architecture \cite{giuliari2020transformer} has been introduced to perform trajectory prediction. Reinforcement learning has also been utilized to generate “safe” trajectories \cite{van2019safecritic}, which could avoid collisions precisely. Pedestrians' social interactions can be captured by a pooling-based \cite{alahi2016social} or graph-based \cite{huang2019stgat} method. Future uncertainties can be handled by generating multi-modal outputs in a conditioned, generative manner \cite{gupta2018social}. The conditions can be sampled from either random Gaussian noise or environmental contexts \cite{syed2019sseg}\cite{haddad2019situation}. However, sampling from environmental contexts drastically increases the computing overhead.

Though remarkable progress has been made over recent years, there are still three shortages which influence the trajectory prediction performance. Firstly, most works encode pedestrians' motion patterns with long short-term memory (LSTM) units \cite{alahi2016social}\cite{gupta2018social}\cite{huang2019stgat}, despite they may attend to later time steps due to the forget gates. Nevertheless, the significant time steps that influence trajectory prediction performance maybe arbitrary time steps in the historical trajectory. It is difficult for a vanilla LSTM to capture such arbitrary but significant time steps for an improved trajectory prediction performance. Secondly, most works capture pedestrians' social interactions in a data-driven manner. However, the obstacle avoidance experience (OAE) which is beneficial to capture pedestrians’ social interactions from limited trajectory data is not fully utilized. Common sense is that other pedestrians in front always influence a pedestrian's future trajectory. How to introduce such common sense into the data-driven learning process is crucial for capturing realistic social interactions. Thirdly, future uncertainties are always handled by generating a latent variable from either random Gaussian noise or observed data. However, pedestrians' future trajectories are highly related to their planning. Several recent works \cite{Song2020PiP}\cite{Mangalam2020It} have tried to generate an informative latent variable from future trajectories. However, such a variable can only be generated in the training stage. Generating an informative latent variable efficiently in the testing stage determines the broad applications of trajectory prediction.

To address the above-discussed limitations in trajectory prediction, we propose an encoder-decoder-based trajectory predictor, named GTPPO (Graph-based Trajectory Predictor with Pseudo Oracle), to forecast pedestrians' trajectories conditioned on their future behaviors. GTPPO fuses information on pedestrians' motion patterns, social interactions, and future behaviors into a concatenated embedding. Then, such an embedding is decoded into future trajectories. Unlike traditional methods that encode pedestrians' motion patterns with LSTMs, a temporal attention-based LSTM has been introduced to improve the encoding process. The improvement is achieved by attending to specific time steps in each pedestrian's historical trajectory, whereas traditional LSTMs may attend to later time steps. Then, a social graph attention module is proposed to capture pedestrians' social interactions. To better aggregate neighbors' information in a realistic manner, OAE is introduced into the data-driven learning process that calculates graph attention. Finally, we regard pedestrians' future behaviors as oracles that guide the trajectory prediction process. We have tried to predict pseudo oracles from  pedestrians' historical behaviors and make them similar to true oracles. Hence, a novel pseudo oracle predictor (POP) module is proposed to generate an informative latent variable which encodes pedestrians' behavior information. Such a variable is beneficial to handle future uncertainties while retaining accurate prediction performance. Specifically, we introduce a Gaussian-LSTM to generate a latent variable from pedestrians' future behaviors in the training stage. Details of the Gaussian-LSTM will be introduced in the section of pseudo oracle predictor. Future behaviors are represented by ground-truth positions, velocities, and accelerations. Meanwhile, we introduce another Gaussian-LSTM to generate a latent variable from observed positions, velocities, and accelerations. The knowledge gap between the two latent variables is minimized during the training stage. Therefore, the latent variable generated from observed data can be used when there is no ground-truth. More importantly, we demonstrate in the experiments that GTPPO can recognize sudden motion changes in the future. Such findings indicate that GTPPO successfully peeks into the future. In conclusion, our main contributions are three-fold as follows:

\begin{itemize}
\item To better encode pedestrians' motion patterns, a temporal attention-based LSTM is introduced as the encoding strategy. Rather than the vanilla LSTM \cite{alahi2016social}\cite{gupta2018social}, the proposed method can highlight critical time steps in each pedestrian's historical trajectory, thus can improve the encoding process. It reveals the importance of critical time steps in historical trajectory, hence indicates a new manner in encoding pedestrians' motion patterns.
\item To capture pedestrians' social interactions, a social graph attention module is proposed to skillfully combine the social attention (OAE) and the graph attention. It is non-trivial to precisely capture social interactions from limited trajectory data with the graph attention only \cite{huang2019stgat}\cite{kosaraju2019social}. However, social attention helps model social interactions among truly relevant pedestrians. Such an attention mechanism could improve the trajectory prediction performance and could be easily embedded into other methods, like SGAN \cite{gupta2018social}.
\item To better handle future uncertainties while retaining accurate prediction, a novel POP module is proposed to generate an informative latent variable based on pedestrians' future behaviors. Compared with generating latent variables from random Gaussian noise \cite{gupta2018social}\cite{sadeghian2019sophie} or time-consuming environmental contexts \cite{syed2019sseg}\cite{haddad2019situation}, the POP module could forecast pedestrians' future behaviors in both training and testing stages and could explore the latent scene structure in a time-saving way by utilizing pedestrians' positions. As far as we know, it is the first attempt to forecast pedestrians' future behaviors from historical trajectories. It could even recognize sudden motion changes in the future, as revealed by qualitative evaluations.
\end{itemize}

Based on the contributions above, we resolve the above-discussed limitations in encoding motion patterns, capturing social interactions, and handling future uncertainties. The computing overhead is low since we do not use image contexts. Finally, we achieve impressive performance on ETH  \cite{pellegrini2010improving},UCY \cite{leal2014learning} and Stanford Drone dataset (SDD) \cite{Robicquet2016Learning} in most cases.

The rest of the paper is organized as follows. Section \uppercase\expandafter{\romannumeral2} reviews related works. Section \uppercase\expandafter{\romannumeral3} describes the proposed method in detail. \uppercase\expandafter{\romannumeral3}.A describes the structure of the encoder-decoder-based generator. \uppercase\expandafter{\romannumeral3}.B introduces the social graph attention module. \uppercase\expandafter{\romannumeral3}.C describes the proposed pseudo oracle predictor and details of the Gaussian-LSTM. Section
\uppercase\expandafter{\romannumeral4} presents the experimental results. Section
\uppercase\expandafter{\romannumeral5} provides a conclusion and discussion.

\section{Related work}
In this section, we provide a brief overview of the existing studies that are close to this work. We point out the differences between the existing studies and our work. Generally, this work aims at tackling the limitations of the current methods mentioned below.
\subsection{Trajectory prediction methods}
Trajectory prediction is a time-series modeling problem that attempts to understand pedestrians' motion patterns. Early researches have focused on predicting future trajectories with the linear, constant velocity, and constant acceleration models \cite{zernetsch2016trajectory}. However, such a simple model cannot understand intricate motion patterns, thus are not suitable for long-term prediction. For longer prediction horizon, flow-based methods \cite{zhi2019spatiotemporal}\cite{molina2018modelling} have been proposed to learn the directional flow from observed trajectories. Subsequently, trajectories are generated by recursively sampling the distribution of future motion derived from the learned directional flow.

To better understand pedestrians' motion patterns, researchers have resorted to several learning-based methods, including Gaussian mixture regression \cite{li2018generic}, Gaussian processes \cite{laugier2011probabilistic}, random tree searching \cite{aoude2011mobile}, hidden Markov models \cite{wang2018learning}, and DBNs \cite{kasper2012object}. Among these methods, DBNs have been commonly used since they can easily incorporate context information \cite{gu2017human}. However, these learning-based methods are nontrivial to handle high-dimensional data. Moreover, it is challenging to design hand-crafted features that are workable in general cases.

The recent rise of deep neural networks provides a new solution to understand pedestrians' motion patterns. Alahi et al. \cite{alahi2016social} presented Social-LSTM, which encodes pedestrians' motion patterns with a shared LSTM. Then, future trajectories were sampled from the encoded embedding in a generative manner. Several GAN-based approaches \cite{gupta2018social}\cite{amirian2019social}\cite{blaiotta2019learning} have been proposed to produce socially acceptable trajectories by adversarially training a generator and discriminator. Context information has been utilized to improve prediction performance. For example, Xue et al. \cite{xue2018ss} and Syed et al. \cite{syed2019sseg} utilized three LSTMs for encoding person, social, and scene scale information and then aggregating them for context-aware trajectory prediction. Ridel et al. \cite{ridel2019scene} presented a joint representation of the scene and past trajectories by using the Conv-LSTM and LSTM, respectively. Lisotto et al. \cite{lisotto2019social} improved Social-LSTM by encompassing prior knowledge about the scene as a navigation map that embodies most frequently crossed areas. Salzmann et al. \cite{salzmann2020trajectron++} proposed Trajectron++, which could achieve state-of-the-art performance. However, its performance is highly related to the processing of input data. Giuliari et al. \cite{giuliari2020transformer} introduced the Transformer structure to perform precise trajectory prediction. However, it suffers from high space-time complexity. Despite the great progress, few works above has further studied the effects of different time steps in encoding motion patterns. Therefore, the information crucial for trajectory prediction at specific time steps may be covered up due to forgetting gates of LSTMs. Here we introduce a temporal attention-based LSTM to encode each pedestrian's motion pattern, aiming at highlighting the specific time steps in historical trajectories.

\subsection{Capturing social interaction}
One challenge of accurate trajectory prediction is how to capture pedestrians' social interactions. Such interactions can be defined by hand-crafted rules, such as social forces \cite{helbing1995social} and stationary crowds' influence \cite{yi2015understanding}. However, such hand-crafted rules model social interactions based only on psychological or physical realization, which alone is insufficient to capture complex crowd interactions. Recent studies have investigated deep learning techniques to capture crowd social interactions. Such studies are mainly divided into pooling-based and graph-based approaches.

For pooling-based approaches, Social-LSTM \cite{alahi2016social} proposed the social pooling layer to aggregate the social hidden state within the local neighborhood of the agent. SGAN \cite{gupta2018social} used a symmetric function to summarize the global crowd interactions. Many other works \cite{amirian2019social}\cite{zhang2019sr}\cite{sadeghian2019sophie} improved the pooling methods with different tricks, aiming at better capturing pedestrians' social interactions.

Different from pooling-based approaches, graph-based approaches formulate the connections between interactive agents with a graph structure. The graph structure is good at capturing the spatial and temporal correlations between different agents \cite{choi2019drogon}\cite{yi2015understanding}, thus, can lead to better trajectory prediction. Moreover, graph models are especially suitable for handling dynamic scenarios \cite{ivanovic2019trajectron} and heterogeneous agents \cite{ma2019trafficpredict}. With the powerful attention mechanism, the graph attention networks \cite{huang2019stgat}\cite{kosaraju2019social} can better capture pedestrians' social interactions.

Despite the advantages above, few works have introduced OAE into capturing pedestrians' social interactions. As reported by Hasan et al. \cite{hasan2018seeing}, knowing the head orientation is beneficial for capturing social interactions. Specifically, pedestrians' future trajectories are always influenced by pedestrians in front. However, it is nontrivial to detect pedestrians' head orientations. Hence, we present a social graph attention module by introducing OAE-based social attention into data-driven learned graph attention. We use pedestrians' velocity orientations to approximate their head orientations.

\subsection{Handling future uncertainties}
Another challenge of accurate trajectory prediction is future uncertainties, which are caused by the inherent multi-modal nature of human paths. Generative model-based methods have gone mainstream since future uncertainties can be handled by generating multi-modal outputs with a latent variable. Many works generate the latent variable from random Gaussian noise \cite{gupta2018social}\cite{sadeghian2019sophie}. However, such a latent variable suffers from lacking information about pedestrians' motion characteristics or scene context.

To generate an informative latent variable, Lee et al. \cite{lee2017desire} fused information on social interactions, scene semantics, and expected reward function. Inspired by \cite{lee2017desire}, some works learned the latent variable from pedestrians' historical trajectories \cite{zhang2019stochastic}\cite{yuan2019diverse} and environmental context \cite{tang2019multiple}. These methods achieve better prediction performance than SGAN. However, it is hard to find generic representations of the environment. Besides, the latent variable learned from historical trajectories has limited information gain compared with the encoded embedding of these trajectories, thus results in limited performance improvement.

\cite{Song2020PiP} proposed to generate the latent variable based on the planning of the ego-vehicle. However, it is hard to obtain pedestrians' future planning or behaviors. Here we propose an alternative to approximate such behaviors with pedestrians' future positions, velocities, and accelerations. Unlike \cite{zhang2019stochastic}\cite{yuan2019diverse}\cite{tang2019multiple}, we generate an informative latent variable by minimizing the knowledge gap between historical observed and ground-truth trajectories. We will show that such a novel latent variable can handle future uncertainties while retaining high prediction performance.

\section{Proposed method}
The trajectory prediction problem is a time-series analysis. For pedestrian \emph{i}, we first term the position ($x_{i}^{t}$, $y_{i}^{t}$) at time step \emph{t} as $p_{i}^{t}$. The aim of trajectory prediction is to estimate the future trajectory $\mathcal{T}_{i}=\left(p_{i}^{T_{obs}+1}, \ldots, p_{i}^{T_{obs}+T_{pred}}\right)$, considering the historical trajectory $\mathcal{H}_{i}=\left(p_{i}^{1}, \ldots, p_{i}^{T_{obs}}\right)$ and pedestrians' social interactions. $T_{obs}$ and $T_{pred}$ are the observed and predicted time horizons, respectively. Then, the trajectory prediction problem is converted into training a parametric model that predicts future trajectory $\mathcal{T}_{i}$ ($\emph{i}$=1,...,$\emph{n}$), which can be formulated as follows:

\begin{equation}
\underset{\Theta}{\arg \max } P_{\theta}\left(\mathcal{T}_{1}, \ldots, \mathcal{T}_{n} | \mathcal{H}_{1}, \ldots, \mathcal{H}_{n}\right),
\end{equation}
where $\Theta$ represents learnable parameters, and $\emph{n}$ represents the number of pedestrians. In general, we predict pedestrians' relative displacements like \cite{alahi2016social}\cite{gupta2018social} in consideration of the generalization performance.

Fig. 2 demonstrates the overview of GTPPO. It consists of three main components, as follows:
\begin{itemize}
\item \emph{Component 1 (Encoder-decoder-based generator)}: We introduce an encoder-decoder-based generator to forecast pedestrians' future trajectories. A shared temporal attention-based LSTM is introduced to encode each pedestrian's motion pattern. The encoded embedding is concatenated with the outputs of components 2 and 3. A vanilla LSTM is then used to decode the concatenated embedding into the relative displacements.
\item \emph{Component 2 (Social graph attention module)}: We present a social graph attention module to capture pedestrians' social interactions. Unlike graph attention network that learns pedestrians' social interactions from trajectory data only, we introduce OAE into the data-driven learning process for an improved prediction performance.
\item \emph{Component 3 (Pseudo oracle predictor)}: We propose a POP module to generate an informative latent variable for handling future uncertainties while retaining accurate trajectory prediction. Such a variable is generated from pedestrians' future behaviors, which are difficult to obtain in practical conditions. Hence, we present an alternative to predict pedestrians' future behaviors from their historical trajectories.
\end{itemize}

\begin{figure*}
\centering
\includegraphics[width=0.8\textwidth]{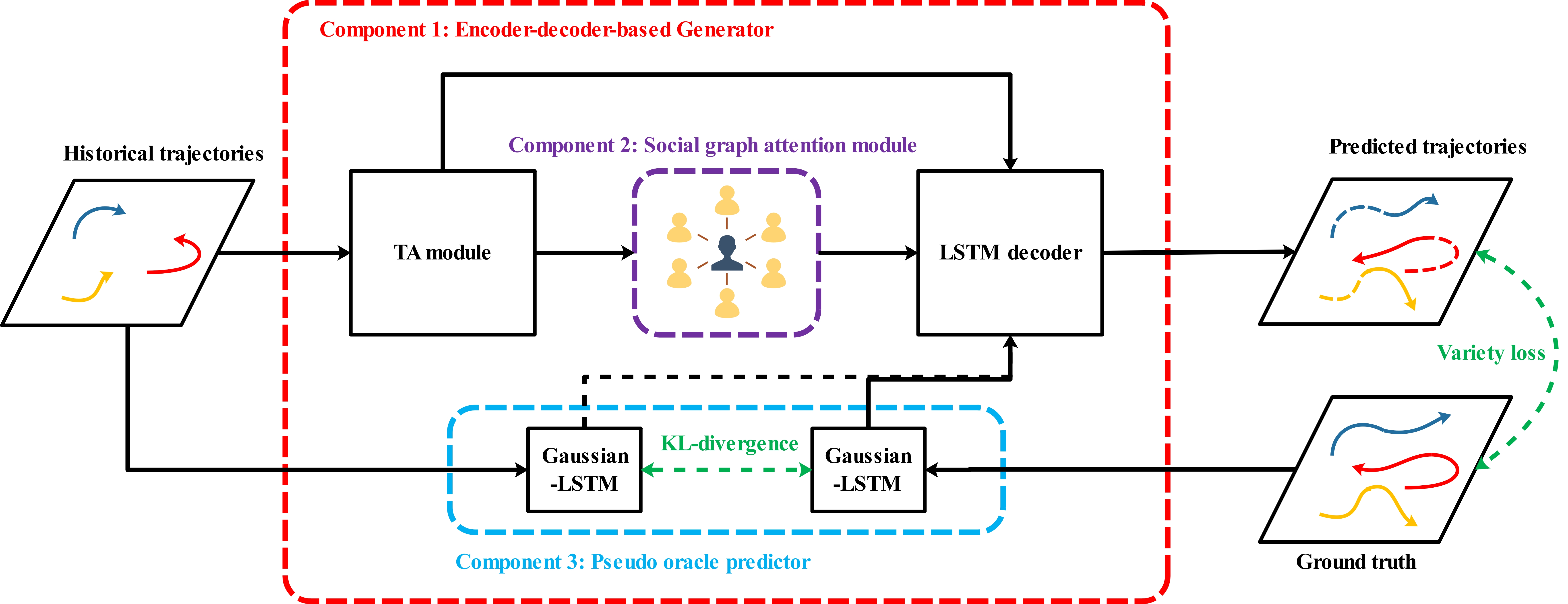}
\caption{The overview of GTPPO. Our model contains three components, including an encoder-decoder-based generator, a social graph module, and a pseudo oracle predictor. Different components are labeled with different colors. The loss functions are green colored (best viewed in color).}
\label{fig:2}
\end{figure*}

\subsection{Encoder-decoder-based generator}
As discussed above, we perform trajectory prediction in a generative manner by introducing an encoder-decoder network. Such a method can generate more realistic trajectories than probabilistic prediction-based methods \cite{lee2017desire}\cite{zhang2019stochastic}, which make Gaussian prior assumption on trajectory generation. We briefly introduce the encoder-decoder structure as follows:

$\textbf{Temporal Attention-based LSTM Encoder}$: Pedestrian's trajectory is a time-series, thus can be modeled by a vanilla LSTM. We introduce a temporal attention mechanism into the vanilla LSTM to highlight specific time steps in observed trajectories. Our method is similar to \cite{Fernando2017Soft}. For brevity, we denote it as a TA module, which works as follows:

First, we use a linear layer to convert the relative displacements of pedestrian \emph{i} at time \emph{t} ($\Delta x_{i}^{t}=x_{i}^{t}-x_{i}^{t-1}$, $\Delta y_{i}^{t}=y_{i}^{t}-y_{i}^{t-1}$) into a fixed-length vector $e_{i}^{t}$. Then, the vector is fed into a vanilla LSTM to encode the embedding of pedestrian \emph{i} at time \emph{t} as follows:

\begin{equation}
e_{i}^{t}=\phi\left(s_{i}^{t} ; W_{e e}\right),
\end{equation}
\begin{equation}
\begin{aligned} m_{i}^{t} =\mathrm{LSTM}\left(m_{i}^{t-1}, e_{i}^{t} ; W_{M}\right), \end{aligned}
\end{equation}
where $\phi(\cdot)$ represents the linear layer function. $m_{i}^{t}$ is the hidden state of the LSTM at time step \emph{t}. $W_{e e}$ and $W_{M}$ are the learnable weights of $\phi(\cdot)$ and the $LSTM(\cdot)$, respectively.

Finally, we further process $m_{i}^{t}$ with a temporal attention mechanism as follows:
\begin{equation}
u_{i}^{t}=\tanh \left(W_{w} m_{i}^{t}+b_{w}\right)
\end{equation}
\begin{equation}
\alpha_{i}^{t}=\frac{\exp \left(u_{i}^{tT} W_{p}\right)}{\sum_{t=1}^{T_{obs}} \exp \left(u_{i}^{tT} W_{p}\right)}
\end{equation}
\begin{equation}
s_{i}=\sum_{t=1}^{T_{obs}} \alpha_{i}^{t} m_{i}^{t}
\end{equation}
where $W_{w}$ and $b_{w}$ are the learnable weight and bias of the activation function $\tanh(\cdot)$. $W_{p}$ is another learnable weight that is learned through continuous training. $s_{i}$ is the output of the TA module.

$\textbf{Vanilla LSTM Decoder}$: We use a vanilla LSTM to decode the concatenated embedding into the relative displacements $(\Delta x_{i}^{t}, \Delta y_{i}^{t})$. The process is as follows:
\begin{equation}
\begin{aligned}
d_{i}^{t+1} &=\mathrm{LSTM}\left(d_{i}^{t}, e_{i}^{t} ; W_{D}\right),
\end{aligned}
\end{equation}
\begin{equation}
\begin{aligned}
\left(\Delta x_{i}^{t+1}, \Delta y_{i}^{t+1}\right) &=\delta\left(d_{i}^{t+1}\right),
\end{aligned}
\end{equation}
where $W_{D}$ is the learnable weight of $LSTM(\cdot)$, $\delta(\cdot)$ is a linear layer that converts the embedding into relative displacements. $d_{i}^{t}$ is the hidden state of the LSTM. Its initialization consists of the output of the TA module $s_{i}$ (defined in Eq.(6)), the output of the social graph attention module $g_{i}^{T_{obs}}$ (defined in Eq.(14)), and the latent variable $z_{i}$ (defined in Section \uppercase\expandafter{\romannumeral3}.C). After getting the predicted relative displacements at each predicted time step, it is easy to convert relative displacements to the future trajectory $\mathcal{T}_{i}$.

\subsection{Social graph attention module}
\begin{figure*}
\centering
\includegraphics[width=0.8\textwidth]{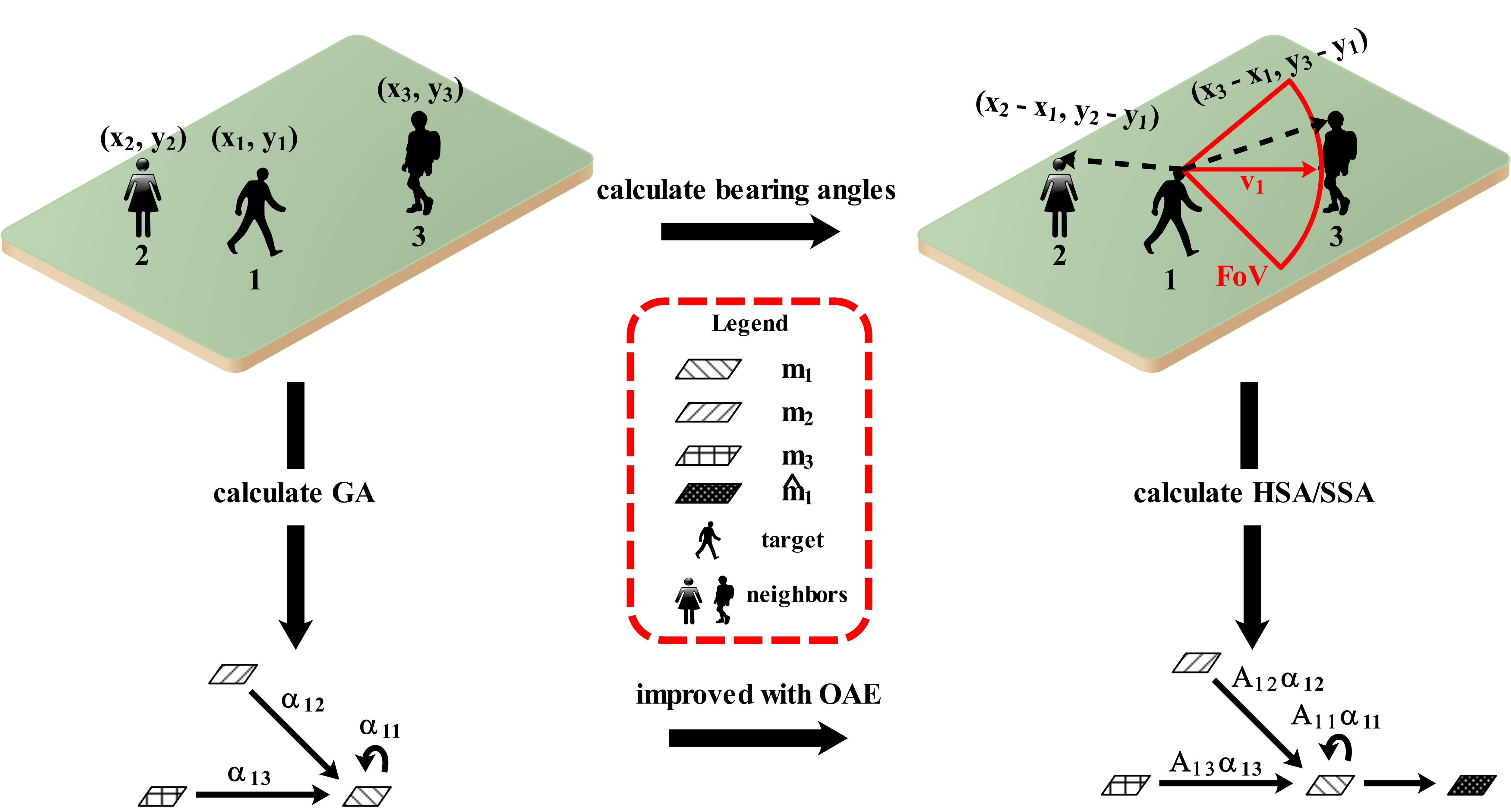}
\caption{Graphical illustration of the social graph attention module. We introduce two attention mechanisms to aggregate neighbors' information. The former is the graph attention $\alpha_{i j}$ learned in a data-driven manner, and the latter is the social attention $A_{i j}$ that is inspired by the OAE that people always influence the future trajectories of pedestrians behind. The social attention is calculated based on pedestrians' velocity orientations.}
\label{fig:3}
\end{figure*}
Pedestrians' social interactions are critical for accurate trajectory prediction. Here we capture such interactions with a graph model due to its ability to model pedestrians and their correlations. As shown in Fig. 3, we fully utilize the data-driven learning power of the graph attention. Besides, we present social attention, which introduces OAE into the data-driven learning process. We briefly introduce them as follows:

$\textbf{Graph attention calculation}$: We introduce a graph attention module, which is similar to the graph attention network \cite{Petar2020Graph} since it allows for aggregating information from neighbors by assigning different importance to different nodes. For brevity, we denote it as GA module. Initially, we feed the encoded embedding $m_{i}^{t}$ (\emph{t}=1,...,$T_{obs}$) (defined in Eq.(3)) to the graph attention layer. Then, we concatenate two transformed embedding $\mathbf{W} m_{i}^{t}$ and $\mathbf{W} m_{j}^{t}$ together to explore their correlation, just like in the Ref \cite{Petar2020Graph}. The attention mechanism is a single-layer feed-forward neural network, parameterized by a weight vector $\mathbf{a}$, and applying the LeakyReLU non-linearity (with input slope 0.2). Finally, the coefficients in the attention layer of the node pair (\emph{i},\emph{j}) at time step \emph{t} are calculated as follows:
\begin{equation}
\alpha_{i j}^{t}=\frac{\exp \left(\operatorname{LeakyReLU}\left(\mathbf{a}^{T}\left[\mathbf{W} m_{i}^{t} \| \mathbf{W} m_{j}^{t}\right]\right)\right)}
{\sum_{k \in \mathscr{N}_{i}} \exp \left(\operatorname{LeakyReLU}\left(\mathbf{a}^{T}\left[\mathbf{W} m_{i}^{t} \| \mathbf{W} m_{k}^{t}\right]\right)\right)}
\end{equation}
where $\|$ is the concatenation operation. $\mathscr{N}_{i}$ represents the neighbors of node \emph{i}. $\mathbf{W}$ and $\mathbf{a}$ are learnable weight matrix and vector, respectively. We concatenate $\mathbf{W} m_{i}^{t}$ and $\mathbf{W} m_{j}^{t}$ together instead of multiplying them through dot products since the latter operation may lead to symmetrical attention coefficients ($\alpha_{i j}^{t}$ = $\alpha_{j i}^{t}$). However, we need asymmetrical attention coefficients ($\alpha_{i j}^{t}$ $\ne$ $\alpha_{j i}^{t}$) since the influence of agent \emph{i} to agent \emph{j} is not equal to that of agent \emph{j} to agent \emph{i}.

After getting the coefficients $\alpha_{i j}^{t}$, the output of one graph attention layer for node \emph{i} at time step \emph{t} is calculated as follows:
\begin{equation}
\hat{m}_{i}^{t}=\sigma\left(\sum_{j \in \mathscr{N}_{i}} \alpha_{i j}^{t} \mathbf{W} m_{j}^{t}\right)
\end{equation}
where $\sigma(\cdot)$ represents the Sigmoid activation function. We stack two graph attention layers in our implementation for stable training.

$\textbf{Social attention calculation}$: The graph's attention captures pedestrians' interactions based on the embedding of their historical trajectories only. However, not all pedestrians truly interact since as common sense, pedestrians' future trajectories are always influenced by people in front of them. \cite{hasan2018seeing} proposed that the pedestrian's head orientation could improve trajectory prediction performance. However, it is nontrivial to calculate the head orientation in wild conditions, even with a state-of-the-art detector \cite{raza2018appearance}. Therefore, we introduce social attention, which attends to pedestrians who are truly interacted based on their velocity orientations. Specifically, we calculate the cosine value of $b_{ij}$, which represents the angle between velocity orientation of agent \emph{i} and the vector joining agents \emph{i} and \emph{j}. All cosine values are formulated as follows:
\begin{equation}
\cos (\mathcal{B})=\left[\begin{array}{ccc}
{\cos \left(b_{1 1}\right)} & {\cdots} & {\cos \left(b_{1 n}\right)} \\
{\vdots} & {\ddots} & {\vdots} \\
{\cos \left(b_{n 1}\right)} & {\cdots} & {\cos \left(b_{n n}\right)}
\end{array}\right],
\end{equation}
where \emph{n} is the number of pedestrians in a scene.

Afterward, we further process $\cos (\mathcal{B})$ with two attention mechanisms as follows:
\begin{itemize}
\item \emph{Hard social attention (HSA):} The proposed hard social attention is similar to the field of view (FOV) filtering, which is performed based on $\cos (\mathcal{B})$. Specifically, we define the  hard attention weight as a matrix $H_{A}$, which has the same size as $\cos (\mathcal{B})$. Each element $h_{ij}$ is set to 1 if $\cos(b_{ij})$ is greater than 0, which is an empirically defined threshold. Otherwise, it is set to 0.
\item \emph{Soft social attention (SSA):} Unlike HSA that calculates attention weights by thresholding, SSA adaptively calculates the attention weights $S_{A}$, which is formulated as follows:
\begin{equation}
S_{A}=\sigma(Conv(\cos (\mathcal{B}))),
\end{equation}
where $\sigma(\cdot)$ represents the Sigmoid activation, and $Conv(\cdot)$ represents the $1 \times 1$ convolutional operation.
\end{itemize}

Then, we can re-write Eq.(10) with social attention weights as follows:
\begin{equation}
\hat{m}_{i}^{t}=\sigma\left(\sum_{j \in \mathscr{N}_{i}} \mathbf{A}_{i j}^{t} \alpha_{i j}^{t} \mathbf{W} m_{j}^{t}\right)
\end{equation}
where $\mathbf{A}$ is the attention weight $H_{A}$ or $S_{A}$.

Finally, we use a vanilla LSTM to process $\hat{m}_{i}^{t}$ since it appears to be a sequence data. We denote it as GLSTM, which is formulated as follows:
\begin{equation}
\begin{aligned}
g_{i}^{t} &=\mathrm{GLSTM}\left(g_{i}^{t-1}, \hat{m}_{i}^{t} ; W_{G}\right),
\end{aligned}
\end{equation}
where $W_{G}$ is the learnable weight of $GLSTM(\cdot)$. $g_{i}^{t}$ is the hidden state of the GLSTM at time step \emph{t}.

\subsection{Pseudo oracle predictor}
\begin{figure*}
\centering
\includegraphics[width=0.8\textwidth]{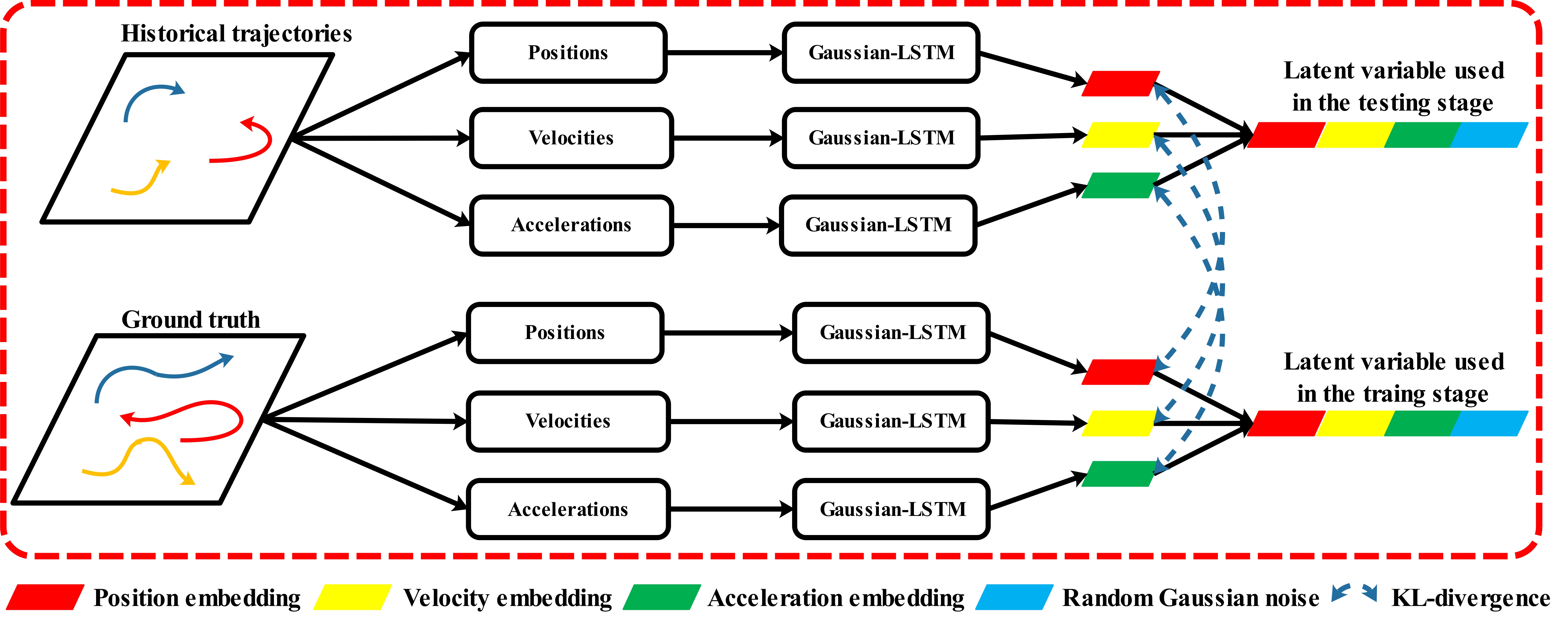}
\caption{Graphical illustration of the POP module. The latent variables generated from ground-truth and observed trajectories are used in the training and testing stages, respectively. KL-divergence is used to minimize the knowledge gap between concatenated embedding of ground-truth and observed trajectories (best viewed in color).}
\label{fig:4}
\end{figure*}
The main highlight of this work is the proposed POP module, which generates an informative latent variable that peeks into pedestrians' future behaviors. Following DESIRE \cite{lee2017desire}, the latent variable is modeled as a Gaussian normal distribution with mean $\mu$ and standard deviation $\sigma$. As shown in Fig. 4, we extract positions, velocities, and accelerations as inputs from historical and ground-truth trajectories, respectively. Except for the velocities that are commonly used in trajectory prediction, positions and accelerations contain potential information about the scene and pedestrians' inherent properties, e.g., anxiety. Then, we encode these inputs by two Gaussian-LSTMs as follows:

\begin{equation}
\left(\mu_{i}^{k}, \sigma_{i}^{k}\right)=F C_{\mu, \sigma}^{k}\left(\Psi^{k}\left(F C_{1}^{k}\left(I_{i}^{k} ; W_{I}^{k}\right) ; W_{L P}^{k}\right) ; W_{\mu}^{k},W_{\sigma}^{k}\right),
\end{equation}
\begin{equation}
\left(\hat{\mu}_{i}^{k}, \hat{\sigma}_{i}^{k}\right)=\hat{FC}_{\mu, \sigma}^{k}\left(\hat{\Psi}^{k}\left(\hat{FC}_{1}^{k}\left(\hat{I}_{i}^{k} ; \hat{W}_{I}^{k}\right) ; \hat{W}_{LP}^{k}\right) ; \hat{W}_{\mu}^{k},\hat{W}_{\sigma}^{k}\right),
\end{equation}
where $\Psi(\cdot)^{k}$ and $\hat{\Psi}(\cdot)^{k}$ are two LSTMs with learnable weights $W_{L P}^{k}$ and $\hat{W}_{L P}^{k}$, respectively. $FC_{1}^{k}$ and $\hat{FC}_{1}^{k}$ are two fully connected (FC) layers that map inputs into 16-dimensional embedding,  with learnable weights $W_{I}^{k}$ and $\hat{W}_{I}^{k}$, respectively. $FC_{\mu, \sigma}^{\mathrm{k}}$ and $\hat{FC}_{\mu, \sigma}^{\mathrm{k}}$ are four FC layers that map outputs of $\Psi(\cdot)^{k}$ and $\hat{\Psi}(\cdot)^{k}$ into four-dimensional latent variable distributions of $\mu_{i}^{\mathrm{k}}$, $\hat{\mu}_{i}^{\mathrm{k}}$, $\sigma_{i}^{\mathrm{k}}$ and $\hat{\sigma}_{i}^{\mathrm{k}}$, with learnable weights $W_{\mu}^{k}$, $\hat{W}_{\mu}^{k}$, $W_{\sigma}^{k}$ and $\hat{W}_{\sigma}^{k}$, respectively. $I_{i}^{k}$ and $\hat{I}_{i}^{k}$ are the $k^{th}$ kind of input (positions, velocities, and accelerations) we extract from observed and ground-truth trajectories, respectively. We introduce KL-divergence to guarantee that $\left(\mu_{i}^{k}, \sigma_{i}^{k}\right)$ is similar to $\left(\hat{\mu}_{i}^{k}, \hat{\sigma}_{i}^{k}\right)$ after training. Therefore, $\left(\hat{\mu}_{i}^{k}, \hat{\sigma}_{i}^{k}\right)$ could be replaced by $\left(\mu_{i}^{k}, \sigma_{i}^{k}\right)$ when ground truth trajectories are unavailable. The generation of the latent variable $z_{i}$ is discrepant in different stages.

$\textbf{In the training stage}$: The estimated latent variable for pedestrian \emph{i} is $z_{i}$, which is generated by concatenating samples from $\left(\hat{\mu}_{i}^{k}, \hat{\sigma}_{i}^{k}\right)$ (\emph{k}=1, 2, 3) and random Gaussian noise. The learned latent variables are beneficial to precise trajectory prediction since they could forecast pedestrians' future behaviors. However, their diversities are constrained by the learned Gaussian distributions, thus cannot provide as much diversity as a random Gaussian noise does. We combine the three learned latent variables with random Gaussian noise to balance accurate and diverse trajectory prediction.

$\textbf{In the testing stage}$: $z_{i}$ is generated by concatenating samples from $\left(\mu_{i}^{k}, \sigma_{i}^{k}\right)$ (\emph{k}=1, 2, 3) and random Gaussian noise. We use the latent variable generated from observed trajectories to approximate that generated from ground-truth trajectories since we have minimized their knowledge gap after well-designed training.

\subsection{Loss function}
The loss function used in this work consists of two parts, namely, the variety and latent variable distribution losses. The variety loss ($\mathcal{L}_{\text {variety}}$) is used to fit the best-predicted trajectory in L2 loss while maintaining diverse outputs. It works as follows: for each pedestrian, the model generates multiple outputs. Then, it chooses the trajectory that has the smallest L2 distance to ground-truth to calculate the variety loss as follows:
\begin{equation}
\mathcal{L}_{\text {variety}}^{i}=\min _{v}\left\|\hat{\mathcal{T}_{i}}-\mathcal{T}_{i}^{v}\right\|_{2},
\end{equation}
where $\hat{\mathcal{T}_{i}}$ and $\mathcal{T}_{i}^{v}$ are ground-truth and the $v^{th}$ predicted trajectories, respectively. $v$ is a hyper-parameter and is set to 20 according to SGAN \cite{gupta2018social}.

The latent variable distribution loss ($\mathcal{L}_{\text {LD}}$) is used to measure the knowledge gap between observed and ground-truth trajectories. We use KL-divergence \cite{goldberger2003efficient} to calculate the loss, which is formulated as follows:
\begin{equation}
\mathcal{L}_{\text {LD}}^{i}=\sum_{k=1}^{3}D_{K L}(\left(\mu_{i}^{k}, \sigma_{i}^{k}\right) || \left(\hat{\mu}_{i}^{k}, \hat{\sigma}_{i}^{k}\right)),
\end{equation}

Afterward, the total loss is defined in a weighted manner as follows:
\begin{equation}
\mathcal{L}_{\text {total}}=\frac{1}{N} \sum_{i=1}^{N} (\mathcal{L}_{\text {variety}}^{i}+\alpha \times \mathcal{L}_{\text {LD}}^{i}),
\end{equation}
where $N$ is the total number of training samples and $\alpha$ is set to 10 by cross-validation across benchmarking datasets.

\subsection{Implementation details}
One layer LSTMs are used for encoder and decoder, in which the dimensions of the hidden states are 32. The 16-dimensional latent variable contains a four-dimensional random Gaussian noise and three four-dimensional vectors that encode positions, velocities, and accelerations, respectively. We train the network with a batch size of 64 for 400 epochs using Adam \cite{kingma2014adam} optimizer. The learning rate of the encoder-decoder network is 0.001, and that of the latent variable predictor is 0.0001. The proposed model is built with the Pytorch framework and is trained with an Intel I7 CPU and an NVIDIA GTX-1080 GPU.

\section{Experimental results}
The proposed method is evaluated on ETH, UCY, and the more challenging SDD, which are publicly available. The former two datasets contain five scenes, namely, ETH, HOTEL, UNIV, ZARA1, and ZARA2. All scenes contain real-world pedestrian trajectories with rich human-human and human-object interaction scenarios, including people crossing each other, groups forming and dispersing, and collision avoidance. All the trajectories of 1,536 pedestrians are converted to real-world coordinates. For ETH and UCY, we use the leave-one-out approach similar to that used by Social-LSTM \cite{alahi2016social}. Specifically, we train models on four scenes and test them on the remaining scene. SDD is a crowd pedestrian dataset with 8 unique scenes on a university campus. Pedestrians' locations are labeled in pixels. We follow the same data-split setting, as used in NMMP \cite{Hu2020Collaborative}. Specifically, we train models on 31 video sequences and test them on 17 video sequences. For all datasets, the observed and predicted horizons are 8 and 12 time steps, respectively. The prediction horizon is denoted as $T_{p r e d}$. More details can be found in Table \uppercase\expandafter{\romannumeral1}. We perform down-sampling to decrease the computing overhead. We only count the number of trajectories whose lengths are equal or greater than 20 time steps (including 8 observed and 12 predicted time steps). Moreover, we generate the training samples by using a sliding time window with a length of 20 and a stride size of one from selected trajectories.

\begin{table}[]
\label{tab1}
\centering
\caption{Details of different datasets.}
\begin{tabular}{lll}
\toprule
Dataset & \begin{tabular}[c]{@{}l@{}}Trajectory number\\ (Length \textgreater{}= 20)\end{tabular} & \begin{tabular}[c]{@{}l@{}}Sampling rate\\ (Hz)\end{tabular} \\
\midrule
ETH    & 44              & 2.5                                                          \\
HOTEL   & 122               & 2.5                                                          \\
UNIV    & 722               & 2.5                                                          \\
ZARA1   & 142               & 2.5                                                          \\
ZARA2   & 189               & 2.5                                                          \\
SDD     & 2829              & 2  \\
\bottomrule
\end{tabular}
\end{table}

Besides, the proposed method is evaluated with two error metrics as follows:

1. \emph{Average Displacement Error (ADE)}: Average L2 distance between the predicted trajectory and the ground-truth trajectory over all predicted horizons.

2. \emph{Final Displacement Error (FDE)}: The Euclidean distance between the predicted and the real final destination at the last predicted step.

\subsection{Quantitative  evaluations}
\textbf{Comparisons with state-of-the-art methods}: Since the commonly used baselines, including the linear regressor, vanilla-LSTM, and social force model perform worse than Social-LSTM \cite{alahi2016social}, we only compare the proposed method against the following state-of-the-art methods:

1. \emph{Social-LSTM} (2016) \cite{alahi2016social}: An improved LSTM-based trajectory prediction method by proposing a social pooling layer to aggregate hidden states of interested pedestrians. Future trajectories are predicted by decoding the concatenation of LSTM embedding and social pooling outputs.

2. \emph{SGAN} (2018) \cite{gupta2018social}: An improved version of Social-LSTM by utilizing adversarial training to generate socially acceptable trajectories. Random Gaussian noises are used as latent variables to generate multi-modal outputs in consideration of pedestrians' future uncertainties.

3. \emph{SR-LSTM} (2019) \cite{zhang2019sr}: An improved version of Social-LSTM by proposing a data-driven state refinement module. Such a module iteratively refines the current pedestrians' hidden states on the basis of their neighbors' intentions through message passing.

4. \emph{Sophie} (2019) \cite{sadeghian2019sophie}: An improved version of SGAN by utilizing attention mechanisms, namely, the social and physical attention modules. The trajectory prediction performance is improved by highlighting the key information with attention operations.

5. \emph{S-Way} (2019) \cite{amirian2019social}: An improved version of SGAN by replacing the L2-loss with the information-loss proposed in Reference \cite{chen2016infogan} to avoid mode collapsing.

6. \emph{STGAT} (2019) \cite{huang2019stgat}: An autoencoder-based trajectory prediction method that uses a spatiotemporal graph attention network to model pedestrians' social interactions in the scene. Specifically, the spatial interactions are captured by the graph attention mechanism, and temporal correlations are modeled by a shared LSTM.

7. \emph{NMMP} (2020) \cite{Hu2020Collaborative}: A trajectory prediction approach that proposes a neural motion message passing strategy to explicitly model the interaction and learn representations for directed interactions between pedestrians.

8. \emph{CVM} (2020) \cite{scholler2020constant}: A simple constant velocity model (CVM) is used to perform trajectory prediction.

9. \emph{Transformer} (2020) \cite{giuliari2020transformer}: A trajectory prediction framework which is designed based on a Transformer network and the larger bidirectional Transformer (BERT) \cite{2018BERT}.

10. \emph{Trajectron++} (2020) \cite{salzmann2020trajectron++}: Trajectron++ is a modular, graph-structured recurrent model that forecasts the trajectories of a general number of agents with distinct semantic classes while incorporating heterogeneous data, e.g., semantic maps and camera images.

\begin{table}[]
\label{tab2}
\centering
\caption{Main properties and differences of the selected state-of-the-art methods for comparisons.}
\begin{tabular}{ll}
\toprule
Social-LSTM(2016) \cite{alahi2016social}  & social pooling layer  \\
SGAN(2018) \cite{gupta2018social}         & adversarial training  \\
SR-LSTM(2019) \cite{zhang2019sr}      & state refinement module  \\
Sophie(2019)  \cite{sadeghian2019sophie}      & social and physical attention  \\
S-Ways(2019) \cite{amirian2019social}       & information-loss \cite{chen2016infogan} \\
NMMP(2020) \cite{Hu2020Collaborative} & neural motion message passing \\
STGAT(2019) \cite{huang2019stgat}        & spatiotemporal graph attention \\
CVM(2020) \cite{scholler2020constant}          & constant velocity model \\
Transformer(2020) \cite{giuliari2020transformer}  & the BERT \cite{2018BERT} structure \\
Trajectron++(2020) \cite{salzmann2020trajectron++} & graph-structured recurrent model \\
\bottomrule
\end{tabular}
\end{table}

Table \uppercase\expandafter{\romannumeral2} sums up the main properties of selected methods. To evaluate different combinations of our methods, we denote GTPPOv1, GTPPOv2, and GTPPOv3 as our methods without and with hard/soft social attention, respectively. Table \uppercase\expandafter{\romannumeral3} presents the comparison results between ours and the state-of-the-art methods. Notably, all methods except Trajectron++ feed pedestrians' relative displacements into encoding modules. Therefore, we reproduce Trajectron++ by replacing their inputs (normalized positions, velocities, and accelerations) with relative displacements for fair comparisons. We also remove the data augmentation strategy used in Trajectron++. We can conclude from Table \uppercase\expandafter{\romannumeral3} as follows:
\begin{itemize}
\item Social-LSTM and SGAN are typical deterministic and generative model-based trajectory predictors that use deep neural networks. However, their performance is not as satisfactory as the recently proposed methods.
\item Both Sophie and S-Ways employ attention mechanisms in capturing social interactions and then achieve an improved prediction performance compared with SGAN. Besides, S-Ways introduces several hand-crafted features, including bearing angles, Euclidean distances, and the future closest distances. Hence, S-Ways achieves competitive performance in the challenging ETH set.
\item SR-LSTM proposes a state refinement module to aggregate neighbors' information. It achieves similar prediction performance as that of S-Ways.
\item NMMP and STGAT use graph models to capture social interactions. Both of them perform better than Social-LSTM, SGAN, and Sophie. The comparisons reveal that the graph model is good at modeling social interactions, which are significant for accurate trajectory prediction.
\item Except for the commonly used LSTM-based trajectory prediction, CVM, and Transformer use a constant velocity model and Transformer structure, respectively. Both of them achieve competitive results in all datasets, which reveal the possibility of performing trajectory prediction without LSTM. Besides, CVM only estimate one output, whereas other methods choose the best output from multiple generated ones.
\item As a recently proposed method, Trajectron++ demonstrates impressive performance. Taking benefit from the usage of the bi-directional GRU encoding module and the graph structure to capture social interactions, Trajectron++ achieves the lowest average ADE value and the second-lowest average FDE value.
\item Our methods, especially GTPPOv3, combine the advantages of the graph model and attention mechanisms. Moreover, an informative latent variable is generated by the proposed POP module. As presented in the table, GTPPOv3 achieves the lowest average FDE value and the second-lowest average ADE value. We owe the impressive performance in average FDE to the POP module since its ability to forecast pedestrians' future behaviors (like short-term destinations) and explore the latent scene structure using trajectory data only
\end{itemize}

\begin{table*}
\label{tab3}
\centering
\caption{ Comparison results with state-of-the-art methods across all datasets. We report ADE and FDE for $T_{p r e d}$ = 12 in meters. Our method preforms impressively by achieving the lowest average FDE value and the second-lowest average ADE value (low is preferred and is labeled with bold fonts). }
\begin{tabular}{p{0.9cm} p{0.9cm} p{0.9cm} p{0.9cm} p{0.9cm} p{0.9cm} p{0.9cm} p{0.9cm} p{0.9cm} p{0.9cm} p{0.9cm} p{0.9cm} p{0.9cm} p{0.9cm} p{0.9cm}}
\toprule
Metric & Dataset & CVM & Social-LSTM & SGAN & SR-LSTM & Sophie & S-Ways & NNMP & STGAT & Trans-former & Plain-Input-Trajectron & Trajec-tron++ & GTPPO-v2 & GTPPO-v3     \\
\midrule
ADE    & ETH     & 0.43 & 1.09        & 0.87 & 0.63    & 0.70   & 0.39  & 0.61         & 0.65 & 0.61 & 0.55 & \textbf{0.35}  & 0.66 & 0.63 \\
       & HOTEL   & 0.19 & 0.79        & 0.67 & 0.37    & 0.76   & 0.39  & 0.33         & 0.35 & \textbf{0.18} & 0.20 & \textbf{0.18}  & 0.35 & 0.19 \\
       & UNIV    & 0.34 & 0.67        & 0.76 & 0.51    & 0.54   & 0.55  & 0.52         & 0.52 & 0.35 & 0.30 &  \textbf{0.22}  & 0.35 & 0.35 \\
       & ZARA1   & 0.24 & 0.47        & 0.35 & 0.41    & 0.30   & 0.44  & 0.32         & 0.34 & 0.22 & 0.20 & \textbf{0.14}  & 0.20 & 0.20 \\
       & ZARA2   & 0.21 & 0.56        & 0.42 & 0.32    & 0.38   & 0.51  & 0.29         & 0.29 & 0.17 & 0.16 & \textbf{0.14}  & 0.26 & 0.18 \\
AVG    &         & 0.28 & 0.72        & 0.61 & 0.45    & 0.54   & 0.46  & 0.41         & 0.43 & 0.31 & 0.28 & \textbf{0.21} & 0.36 & 0.31 \\
FDE    & ETH     & 0.80 & 2.35        & 1.62 & 1.25    & 1.43   & \textbf{0.64}  & 1.08         & 1.12 & 1.12 & 1.08 & 0.77  & 1.01 & 0.98 \\
       & HOTEL   & 0.35 & 1.76        & 1.37 & 0.74    & 1.67   & 0.66  & 0.63         & 0.66 & \textbf{0.30} & 0.35 & 0.38  & 0.59 & \textbf{0.30} \\
       & UNIV    & 0.71 & 1.40        & 1.52 & 1.10    & 1.24   & 1.31  & 1.11         & 1.10 & 0.60 & 0.61 & \textbf{0.48}  & 0.60 & 0.60 \\
       & ZARA1   & 0.48 & 1.00        & 0.68 & 0.90    & 0.63   & 0.64  & 0.66         & 0.69 & 0.38 & 0.38 & \textbf{0.28}  & 0.35 & 0.32 \\
       & ZARA2   & 0.45 & 1.17        & 0.84 & 0.70    & 0.78   & 0.92  & 0.61         & 0.60 & 0.32 & 0.31 & \textbf{0.30}  & 0.37 & 0.31 \\
AVG    &         & 0.56 & 1.54        & 1.21 & 0.94    & 1.15   & 0.83  & 0.82         & 0.83 & 0.55 & 0.55 & \textbf{0.45}  & 0.58 & 0.50 \\
\bottomrule
\end{tabular}
\end{table*}

Table \uppercase\expandafter{\romannumeral4} presents the comparison between the proposed GTPPO-v3 against state-of-the-art methods on SDD. The results of Social-LSTM, SGAN, and Sophie are reported in NNMP \cite{Hu2020Collaborative}. Results of CVM, STGAT, Transformer, and Trajectron++ are obtained by reproducing the publicly released codes following details presented in their original works. Notably, we found Trajectron++’s prediction performance is highly related to the processing of input data. Hence, we also report the results of a Plain-Input-Trajectron, which adopts the same input data as other methods for fair comparisons. We can conclude from Table \uppercase\expandafter{\romannumeral4} as follows:
\begin{itemize}
\item CVM has the worst performance on SDD since the simple constant velocity model maybe not suitable for predicting pedestrians' trajectories in challenging scenes.
\item Social-LSTM, SGAN, Sophie, NNMP, and STGAT achieve better performance on SDD compared with CVM. Among them, NNMP and STGAT achieve competitive performance due to the introduction of graph-based models to capture social interactions.
\item Transformer, Trajectron++, Plain-Input-Trajectron, and GTPPO-v3 achieve impressive performance on SDD due to tricks like the Transformer structure, the bi-directional GRU encoding, and the POP module, respectively. Among them, Trajectron++ achieves the best performance, however, its prediction performance is highly related to the processing of input data. By using the same input data, GTPPO-v3 is slightly superior to Transformer and Plain-Input-Trajectron. A probable reason may be that the POP module could learn pedestrians' future behaviors and explore the latent scene structure in the testing stage. Learning these informative factors help our model improve the trajectory prediction performance. However, Transformer and Plain-Input-Trajectron do not explore semantic information.
\end{itemize}

\begin{table*}
\label{tab4}
\centering
\caption{ Comparison results with state-of-the-art methods on SDD. We report ADE and FDE for $T_{p r e d}$ = 12 in pixels. Our method outperforms others in terms of average ADE and FDE (low is preferred and is labeled with bold fonts). }
\begin{tabular}{p{0.9cm} p{0.9cm} p{0.9cm} p{0.9cm} p{0.9cm} p{0.9cm} p{0.9cm} p{0.9cm} p{0.9cm} p{0.9cm} p{0.9cm}}
\toprule
Metric & CVM & Social-LSTM & SGAN        & Sophie   &  NNMP   & STGAT & Trans-former & Plain-Input-Trajectron & Trajec-tron++ & GTPPO-V3 \\
\midrule
ADE  &  175.54    & 31.19 & 27.25 & 16.27    &  14.67   &  12.90     &  13.09      &   12.18  &  11.92 & \textbf{10.13}     \\
FDE  &  290.54    & 56.97 & 41.44 & 29.38    &  26.72   &  24.72     &  17.13      &   18.02  &  16.42 & \textbf{15.35}     \\
\bottomrule
\end{tabular}
\end{table*}

\textbf{Ablation study}:
We perform an ablation study to investigate the effects of different modules used in GTPPO. We use a one-layer LSTM-based encoder-decoder network as the baseline. We can obtain three conclusions from Table \uppercase\expandafter{\romannumeral5}, as follows:
\begin{itemize}
\item Results in the $1^{st}$ and $2^{nd}$ rows indicate that the proposed TA module is better than a vanilla LSTM in encoding pedestrians' motion patterns, since the former can highlight significant time steps in historical trajectories in a data-driven manner.
\item Results in the $3^{rd}$, $4^{th}$, and $5^{th}$ rows indicate that the GA module can improve the prediction performance by capturing pedestrians' interactions. Besides, the introduced social attention can further improve the prediction performance by introducing OAE into the graph attention. Further, the SSA performs better than the HSA since the former models pedestrians' interactions in a flexible way, whereas the latter models those in a deterministic manner.
\item Comparisons between the $1^{st}$ and $6^{th}$ rows verify the effect of the proposed POP module in generating an informative latent variable, which is critical for accurate trajectory prediction. Specifically, the generated latent variable encourages the model to explore the knowledge about pedestrians' future trajectories.
\end{itemize}

More details can be found in Table \uppercase\expandafter{\romannumeral5}. In short, the combination of TA, GA, SSA, and POP achieves the best performance in means of average ADE and FDE. The ablation study verifies the effects of our contributions.

\begin{table*}
\label{tab5}
\centering
\caption{ Ablation study. We report average ADE and FDE for $T_{p r e d}$ = 12 in meters across five datasets. TA represents the temporal attention-based LSTM. GA represents the graph attention. HSA and SSA represent hard and soft social attention, respectively. POP represents the pseudo oracle predictor (low is preferred and is labeled with bold fonts). }
\begin{tabular}{llllllllll}
\toprule
Row & TA & GA & HSA & SSA & POP & average ADE & average FDE \\
\midrule
 1& $\times$       &   $\times$       &    $\times$      &      $\times$        &    $\times$                 & 0.56           & 1.19           \\
 2&\checkmark         &   $\times$       &  $\times$        &  $\times$        &   $\times$           & 0.54           & 1.13           \\
 3& $\times$          & \checkmark               &  $\times$            &   $\times$                   &  $\times$         & 0.46           & 0.86           \\
 4& $\times$           &  \checkmark                & \checkmark         &  $\times$     &      $\times$      & 0.42           & 0.82           \\
 5& $\times$                  &   \checkmark      &      $\times$     &  \checkmark       &      $\times$      & 0.39           & 0.75           \\
 6&$\times$                   &    $\times$             &    $\times$      &    $\times$          & \checkmark        & 0.42           & 0.72           \\
 7&\checkmark       & \checkmark               &   $\times$         &    $\times$        &     $\times$                    & 0.43           & 0.81           \\
 8&\checkmark           & \checkmark     &  \checkmark          &  $\times$      &    $\times$                     & 0.41           & 0.76           \\
 9&\checkmark       & \checkmark     &   $\times$        & \checkmark               &    $\times$                     & 0.39           & 0.65           \\
 10&\checkmark                  &  $\times$     &   $\times$       &    $\times$         & \checkmark                       & 0.40           & 0.69           \\
 11&$\times$       & \checkmark               &   $\times$      &     $\times$          & \checkmark                       & 0.39           & 0.61           \\
 12&$\times$        & \checkmark             &  \checkmark       &     $\times$          & \checkmark                       & 0.39           & 0.61           \\
 13&$\times$       &  \checkmark      &    $\times$     & \checkmark                 & \checkmark                       & 0.34           & 0.54           \\
 14&\checkmark      & \checkmark               &   $\times$          &    $\times$       & \checkmark                       & 0.38           & 0.60           \\
 15&\checkmark        &  \checkmark       & \checkmark        &     $\times$             & \checkmark                       & 0.36           & 0.58         \\
 16&\checkmark      & \checkmark        &    $\times$    &  \checkmark   & \checkmark                & \textbf{0.31}           & \textbf{0.50}           \\
\bottomrule
\end{tabular}
\end{table*}

\textbf{Evaluations of different sampling numbers}:
A generative model-based trajectory predictor handles future uncertainties by generating multiple outputs. However, many outputs are far away from the ground-truth, thus may influence or mislead further decisions which are based on the predicted trajectories. We claim that a good trajectory predictor should estimate accurate future trajectories with few attempts while maintaining diverse outputs. As discussed before, the POP module encourages the model to investigate pedestrians' future behaviors. Hence, we assume that POP is beneficial to accurate trajectory prediction with few attempts.

To verify our assumption, we introduce STGAT-POP, which is a modified version of STGAT. Specifically, we replace STGAT's latent variable generated from random Gaussian noise with an informative one generated by the POP module. Fig. 5 illustrates the comparison results of average ADE and FDE between STGAT and STGAT-POP when using different sampling numbers. For both methods, the prediction performance goes worse when gradually decreasing sampling numbers. However, STGAT-POP can still perform satisfactory trajectory prediction with few samplings. As demonstrated in Fig. 5, the average ADE of STGAT-POP with five samplings is 0.43, whereas that of STGAT with 20 samplings is 0.45. The average FDE of STGAT-POP with one sampling is better than that of STGAT with 20 samplings. All these findings indicates that the POP module can predict accurate trajectories with few attempts. Moreover, the POP module can be easily integrated into generative model-based trajectory prediction methods for an improved prediction performance.

\begin{figure}
\centering
\includegraphics[width=0.4\textwidth]{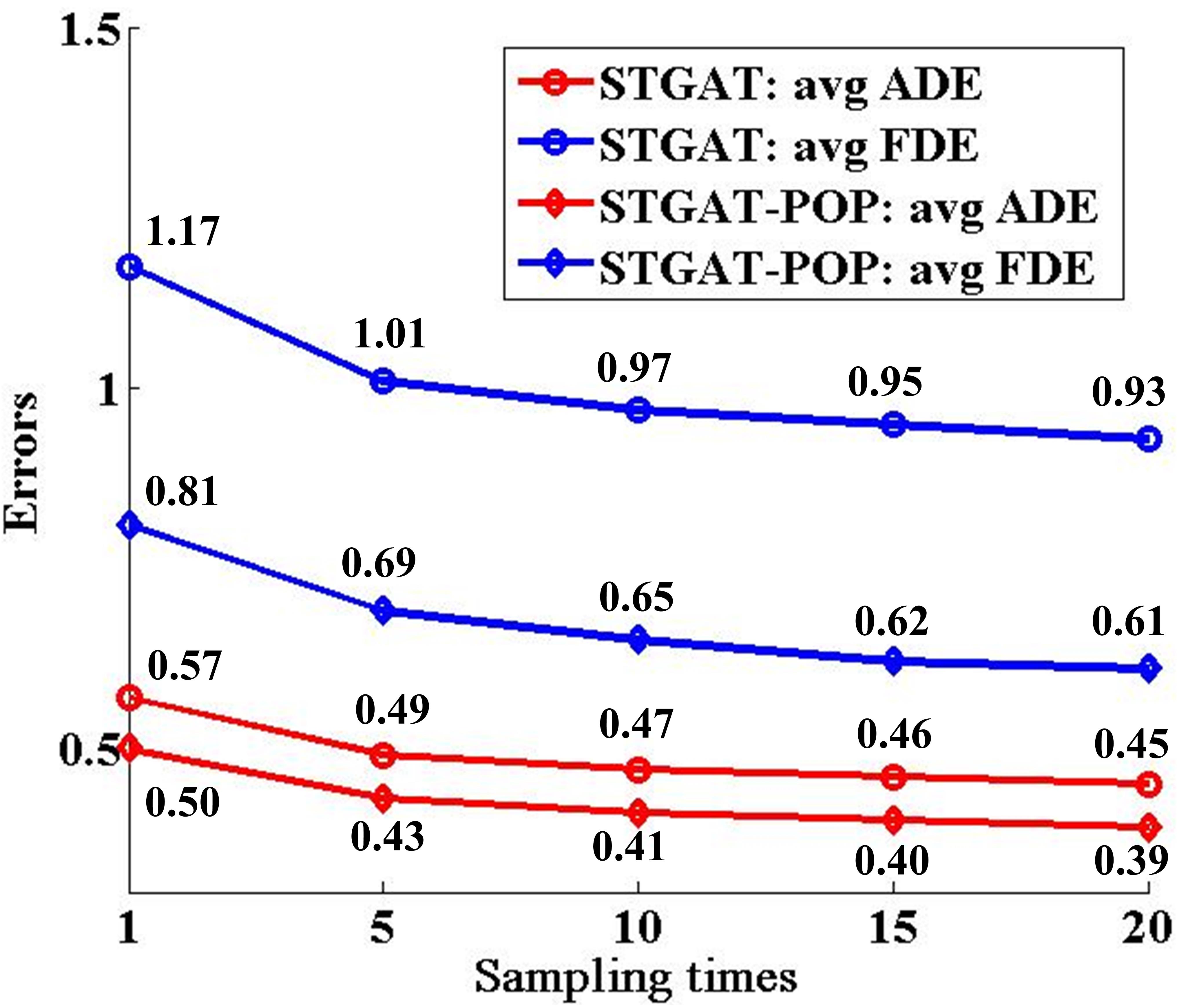}
\caption{Comparisons of average ADE (red lines) and FDE (blue lines) values between STGAT and STGAT-POP when using different sampling numbers across all datasets. The diamond marker represents STGAT-POP, and the circle marker represents STGAT (best viewed in color.)}
\label{fig:5}
\end{figure}

\textbf{Illustrations of the TA module}:
As we claimed before, the vanilla LSTM may attend to later time steps, while the proposed TA module can attend to specific time steps. To evaluate the TA module, we introduce the baseline (a vanilla LSTM-based network) used in the ablation study for comparison. Yellow and black dotted lines represent the prediction results generated by the TA module and the vanilla LSTM-based network, respectively. These two samples are chosen from the UNIV scene. Different attention weights are represented by circles with different sizes, and weight values are labeled around circles. Notably, the sizes of circles are not precisely proportional to their weight values since gaps among different weight values maybe too large. Fig. 6(a) shows the example of the superiority of the TA module compared with the vanilla LSTM-based network. It is obvious that the encoding process of the TA module attends to the last four time steps, which are useful to infer pedestrians' movement trends. Fig. 6(b) shows the example where two methods achieve similar performance. As shown in the figure, the encoding process of the TA module mainly attends to the last two time steps. Such a finding may verify our hypothesis (the vanilla LSTM may attend to later time steps) if we assume that similar performance is obtained by similar encoding processes.

\begin{figure*}
\centering
\includegraphics[width=0.8\textwidth]{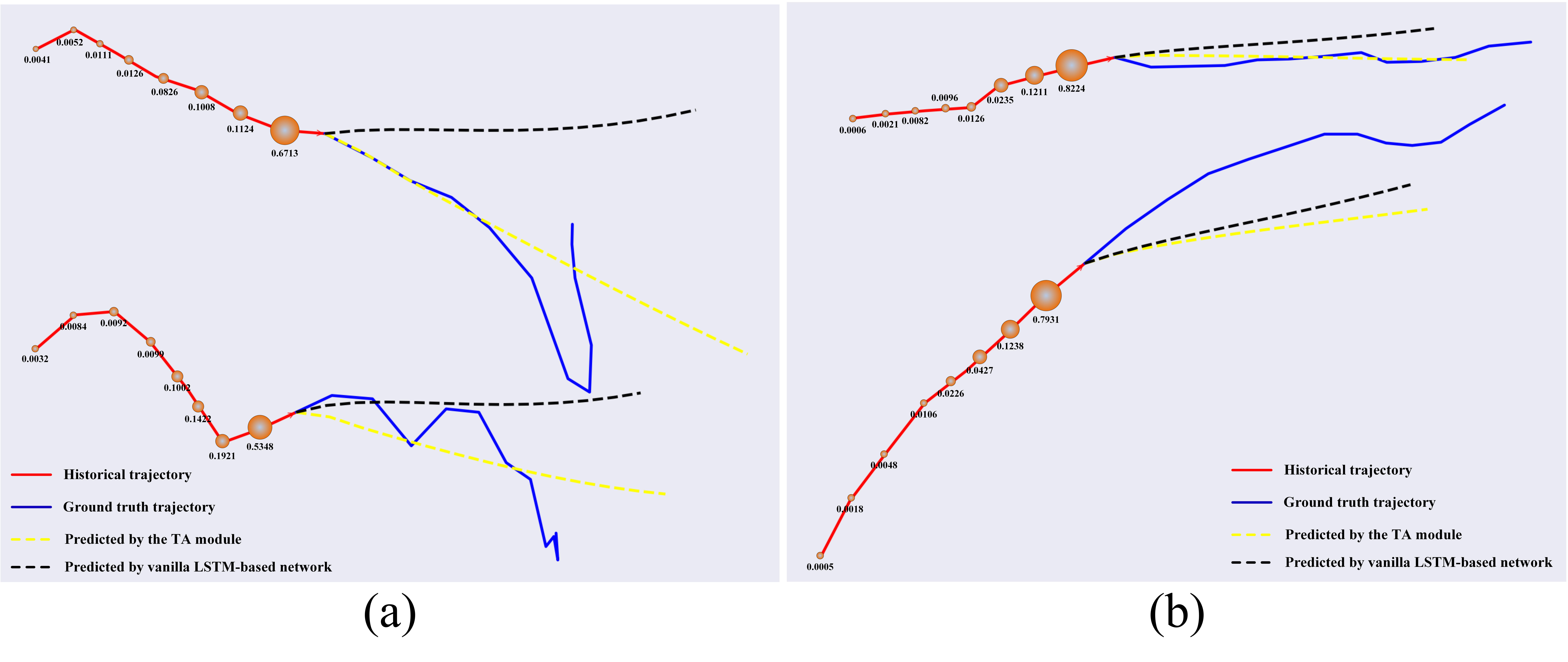}
\caption{Illustrations of different attention weights generated by the TA module. An one-layer LSTM-based encoder-decoder network is introduced for comparison. Different attention weights are represented by circles with different sizes and weight values are labeled around circles (best viewed in color).}
\label{fig:6}
\end{figure*}

\textbf{Computational time analysis}:
The computational time of trajectory prediction is a critical issue for real-time applications. Therefore, we compare the computational times between the proposed methods and two state-of-the-arts (SGAN and STGAT) on the computing platform with an Intel I7 CPU and an NVIDIA GTX-1080 GPU. As presented in Table \uppercase\expandafter{\romannumeral6}, GTPPOv1 takes similar computational times as that of STGAT. Besides, both methods need about two times of computing overheads compared with SGAN since the calculation of graph attention at each time step. GTPPOv2 and GTPPOv3 need about four times of computing overheads compared with SGAN due to the calculation of the social attention. Notably, we simultaneously process 64 scenes in once forward operation. Therefore, the proposed method can satisfy the needs of real-time applications. Moreover, GTPPOv1 is a suitable alternative for better real-time performance in scenes that are not so crowded.

\begin{table*}
\label{tab6}
\centering
\caption{ Computational times of the proposed methods compared with STGAT and SGAN. We calculate the computational times for once forward operation. The unit of the time is millisecond.}
\begin{tabular}{llllll}
\toprule
Method & SGAN & STGAT & GTPPOv1 & GTPPOv2 & GTPPOv3 \\
\midrule
Time   & 20 & 42( $\approx$2.0X)    & 45( $\approx$2.0X)  & 82( $\approx$4.0X)      & 84( $\approx$4.0X)  \\
\bottomrule
\end{tabular}
\end{table*}

\subsection{Qualitative evaluations}
We perform qualitative evaluations to explore GTPPO further. Fig. 7 demonstrates the best trajectories generated by STGAT, SGAN, NMMP, Trajectron++, GTPPOv1, GTPPOv2, and GTPPOv3 in different datasets. The best trajectory is the one that has the lowest ADE value among the 20 samples generated by each method. Each sub-figure comprises four scenarios chosen from ETH, HOTEL, ZARA1, ZARA2, and SDD. Generally, all methods could precisely forecast pedestrians' future trajectories when they show linear motions, e.g., target one in the $4^{th}$ scenario of Fig. 7(b). In this scenario, all methods successfully recognize the other two still pedestrians. However, some methods perform struggling when pedestrians exhibit intricate motion patterns. We can conclude the superiorities of our method from Fig. 7 as follows:
 \begin{itemize}
\item Trajectron++ could predict trajectories most close to the ground truth trajectories in ETH, HOTEL, ZARA1, and ZARA2. However, GTPPOv3 could predict short-term destinations closer to the ground truth destination when pedestrians exhibit sudden motion changes in the future (for example, target one in the $1^{st}$ scenario of Fig. 7(d)). Such performance complies with the comparison results in Table \uppercase\expandafter{\romannumeral3}. GTPPOv3's ability in predicting short-term destinations benefits from the POP module, which could forecast pedestrians' future behaviors from observed trajectory data. In Fig. 7(e), GTPPOv3 outperforms others by generating more precise trajectories.
\item As shown in the $2^{nd}$ scenario of Fig. 7(d), target three's future trajectories generated by GTPPOv2 and GTPPOv3 exhibit avoidance since targets one and two are in front. However, target three's future trajectory generated by GTPPOv1 does not exhibit avoidance. Such a difference indicates the effects of the social graph attention module in capturing pedestrians' social interactions.
\end{itemize}

Despite the superiorities mentioned above, some drawbacks are also exposed in Fig. 7. Two significant drawbacks are listed as follows:
 \begin{itemize}
\item As shown in the figure, Trajectron++ could generate more complex motion patterns than other methods. We guess it is the reason why Trajectron++ achieves the lowest average ADE in ETH and UCY sets. One of our future works is how to generate more complex motion patterns.
\item As shown in the $4^{th}$ scenario of Fig. 7(d), all methods fail in predicting future trajectories of targets one to three. The POP module could not predict these targets' short-term destinations since it gains little information from their short historical movements. Such a fail reveals that it is challenging for our method to forecast still pedestrians' future trajectories.
\end{itemize}

\begin{figure*}
\centering
\includegraphics[width=1.0\textwidth]{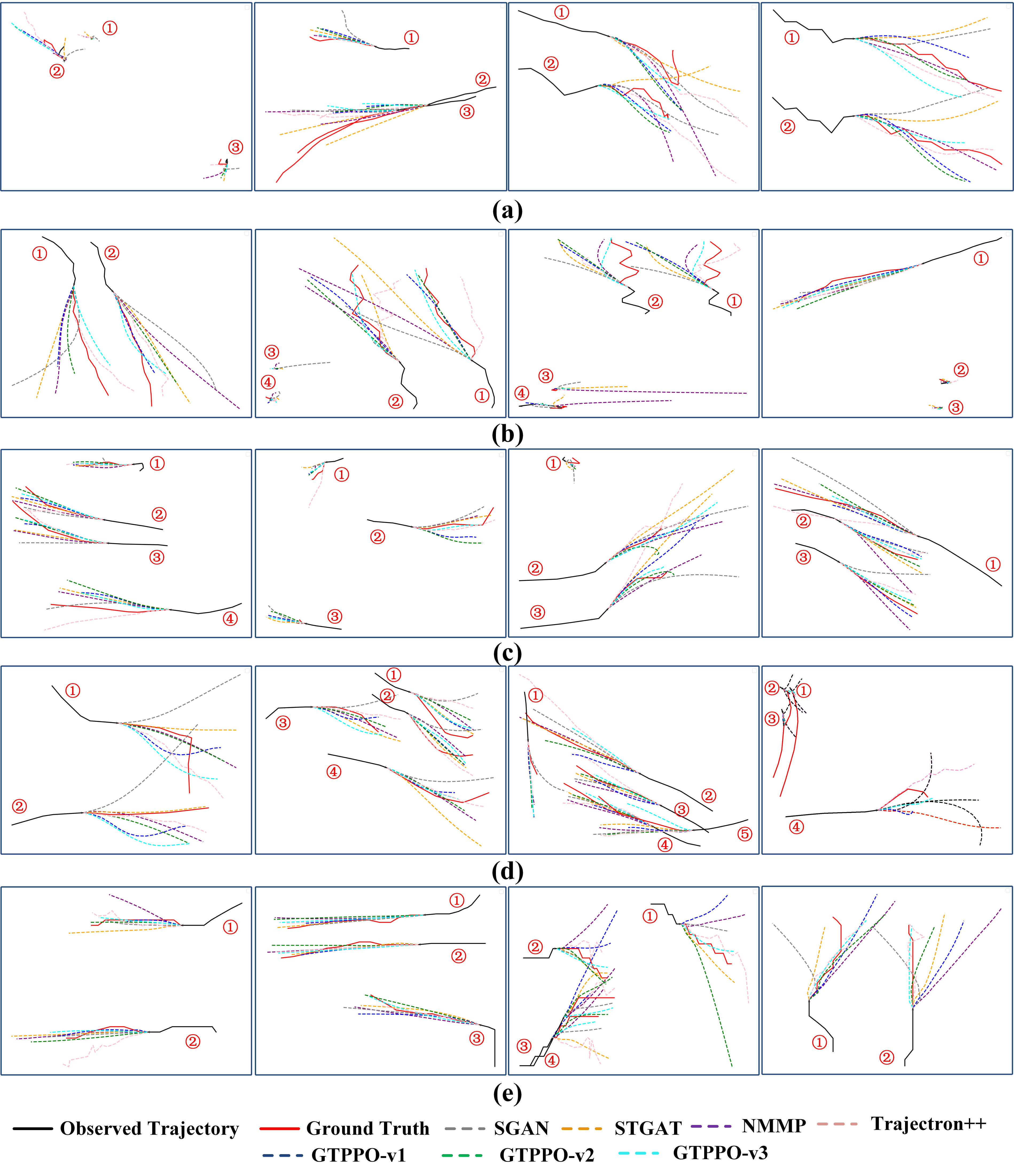}
\caption{Best Trajectories generated by STGAT, SGAN, NMMP, Trajectron++, GTPPOv1, GTPPOv2, and GTPPOv3 in (a) ETH, (b) HOTEL, (c) ZARA1, (d) ZARA2 and (e) SDD datasets. Black and red lines represent observed and ground-truth trajectories, respectively. Dashed lines of different colors represent trajectories predicted by different methods (best viewed in color and zoom-in).}
\label{fig:7}
\end{figure*}

Except for best trajectories, we also compare the density maps generated by STGAT, NMMP, Trajectron++, and GTPPOv3 to evaluate their trajectory prediction performance. Density maps are generated by repeated sampling 300 times from different models. We do not use GTPPOv2 since it is inferior to GTPPOv3 as revealed in Table \uppercase\expandafter{\romannumeral3}. Fig. 8 illustrates the density maps of different methods in six scenarios selected from (a) ETH, (b) HOTEL, (c) ZARA1, (d) ZARA2, (e) COUPA, and (f) HYANG sets. Density maps in the UNIV set are not exhibited since there are too many trajectories in each scenario. Generally, density maps generated by different methods reflect the distributions of pedestrians' future trajectories in most cases. However, GTPPOv3 exhibits better performance in handling pedestrians' sudden motion changes and social interactions. We can conclude GTPPOv3's superiorities as follows:
 \begin{itemize}
\item Compared with other methods, density maps generated by GTPPOv3 are closer around the ground truth trajectories (e.g., target one in Fig. 8(c) and target two in Fig. 8(e)). Such a difference further verifies GTPPOv3's ability in forecasting pedestrians' short-term destinations owe to the proposed POP module. More importantly, GTPPOv3 could predict sudden motion changes in many scenarios, e.g., target one in Figs. 8(d) and 8(e), target two in Fig. 8(a), and target three in Fig. 8(f). Such an ability is owed to the POP module, which could forecast pedestrians' future behaviors and explore latent scene structures.
\item In the HOTEL scenario, GTPPOv3 generates more separate density maps for three targets, whereas other methods generate tangly density maps, which may lead to collisions. Such a difference verifies GTPPOv3's ability to capture pedestrians' social interactions better and avoid collisions due to the introduced OAE.
\end{itemize}

Generally, GTPPOv3 could predict accurate and diverse outputs in a socially acceptable way.

\begin{figure*}
\centering
\includegraphics[width=1.0\textwidth]{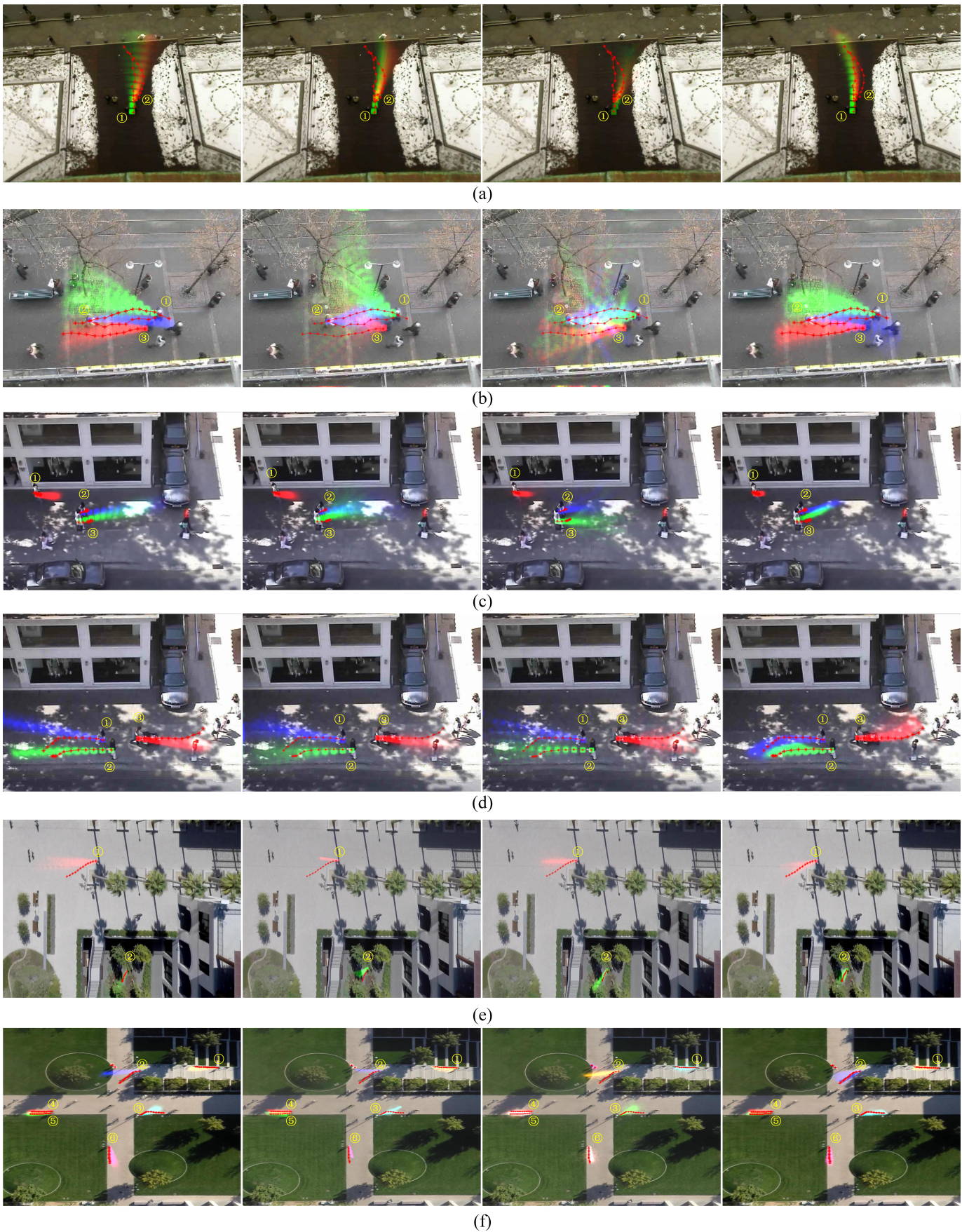}
\caption{Density maps of the predicted trajectories in (a) ETH, (b) HOTEL, (c) ZARA1, (d) ZARA2, (e) COUPA, and (f) HYANG sets. The four columns represent density maps generated by STGAT, NMMP, Trajectron++, and GTPPOv3, respectively. The density maps are generated by sampling 300 times from the well-learned generators. The red stars represent the ground-truth future trajectories, and different colors indicate the density distributions of differently annotated pedestrians (best viewed in color and zoom-in).}
\label{fig:8}
\end{figure*}

\section{Conclusion and discussion}
In this work, we propose GTPPO, which forecasts pedestrians' future trajectories with an encoder-decoder network. For each pedestrian, a TA module is used to encode the historical trajectory, aiming at highlighting the informative time steps. The social interactions between different pedestrians are captured by a social graph attention module, which introduces OAE into graph attention for improved prediction performance. Moreover, a novel POP module is proposed to generate an informative latent variable, which handles future uncertainties by peeking into pedestrians' future behaviors. As revealed by the evaluation results, S-Ways \cite{amirian2019social} achieves competitive results on the challenging ETH scene due to the use of multiple hand-designed interaction features. Trajectron++ \cite{salzmann2020trajectron++} achieves the state-of-the-art performance on the rest scenes of ETH/UCY since the used semantic maps and image information. The proposed GTPPO achieves the second-best performance on ETH/UCY, and the best performance on the challenging SDD.

This approach is promising as a robust trajectory prediction model to avoid pedestrian-vehicle collisions in complex traffic environments. Also, it can help architects to improve their design for emergent evacuation in crowded public areas. For the potential problems of missing and noisy data in robotic applications, the median filtering and data interpolation could be introduced to handle such problems. We will release the code in the future to promote further researches and applications. However, as shown in Table \uppercase\expandafter{\romannumeral3}, Trajectron++ is superior to the proposed method due to the introduced semantic maps. Therefore, one of our future works focuses on generating semantic maps of different scenes with graph neural networks. Another potential improvement is to learn how pedestrians plan their future trajectories with inverse reinforcement learning.

\section*{Acknowledgment}
This work has been supported by the Chang zhou Application Foundation Research Project under Grant CJ20200083, by the Postgraduate Research Practice Innovation Program of Jiangsu Province under Grant YPC20020175,  in part by the National Natural Science Foundation of China under Grant 61801227, and by the Qing Lan Project of Jiangsu Province under Grant QLGC2020.


%




\ifCLASSOPTIONcaptionsoff
  \newpage
\fi



%


\bibliographystyle{IEEEtran}
\bibliography{bib}

%

\begin{IEEEbiography}[{\includegraphics[width=1in,height=1.25in,clip,keepaspectratio]{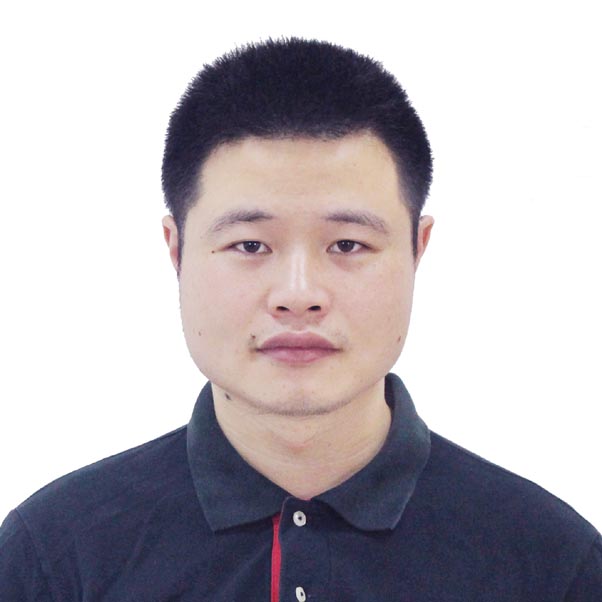}}]{Biao Yang}
received his BS degree from Nanjing University of Technology. He received his MS and Ph.D degrees in instrument science and technology from Southeast University, Nanjing, China, in 2014. From 2018 to 2019 he was a visiting scholar in the University of California, Berkeley. Now he works at Changzhou University, China. His current research interests include machine learning and pattern recognition based on computer vision.
\end{IEEEbiography}
\vspace{-35pt}
\begin{IEEEbiography}[{\includegraphics[width=1in,height=1.25in,clip,keepaspectratio]{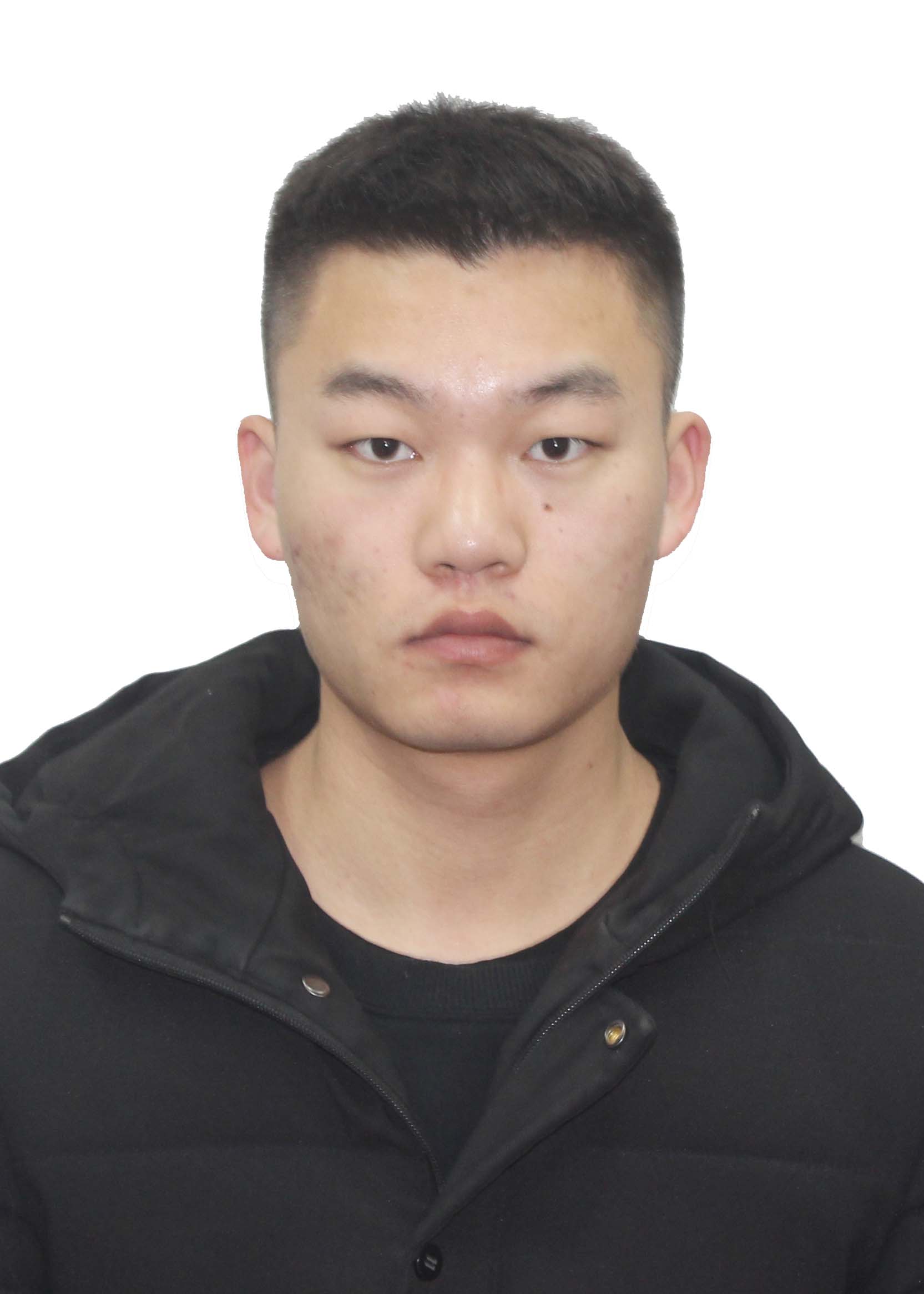}}]{Guocheng Yan}
was born in Fuyang, china, in 1996. He received the B.S degree in applied physics from Fuyang Normal University 2019. He is currently pursuing the master degree in machine learning.
\end{IEEEbiography}
\vspace{-35pt}
\begin{IEEEbiography}[{\includegraphics[width=1in,height=1.25in,clip,keepaspectratio]{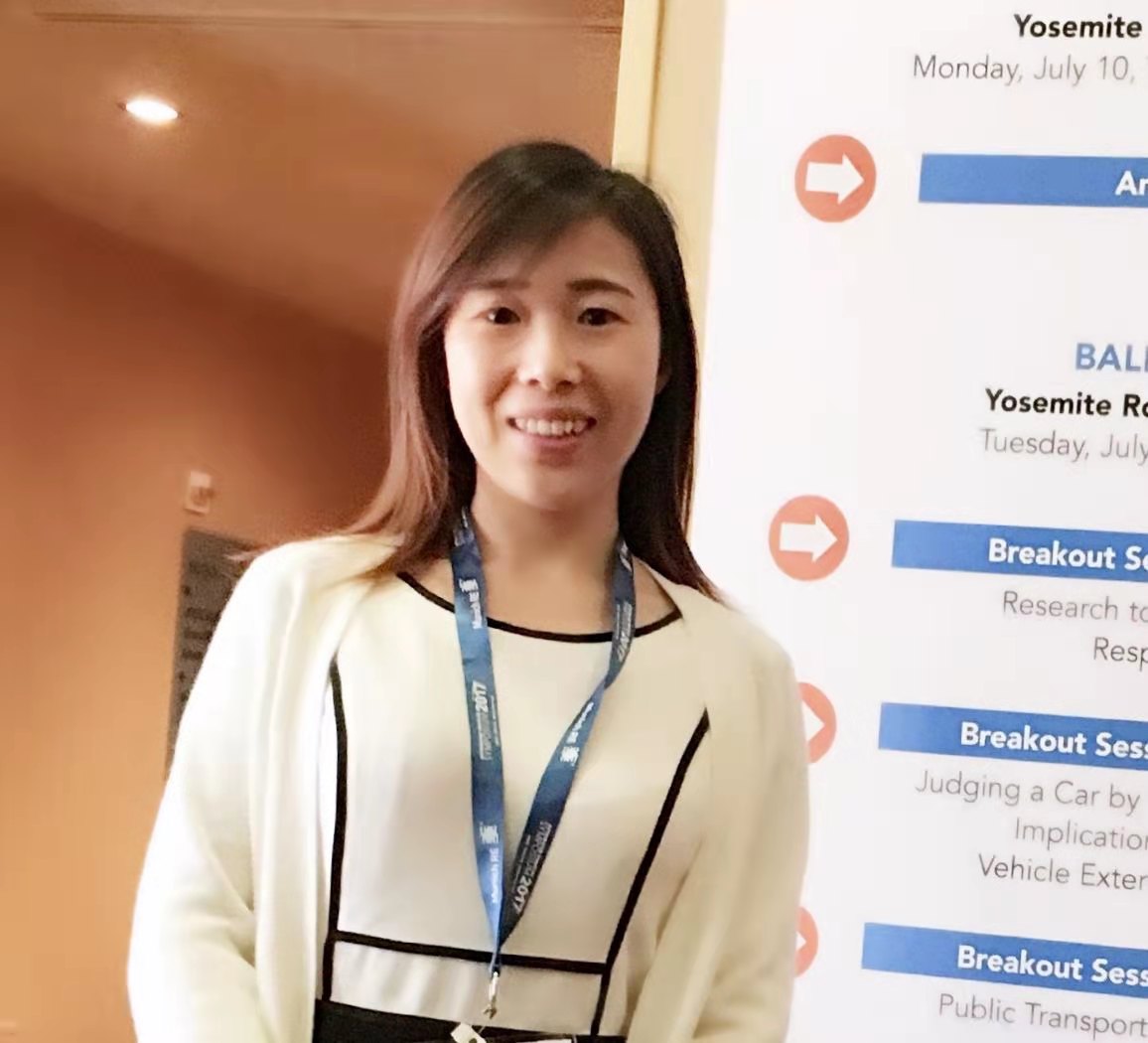}}]{Pin Wang}
received the Ph.D. degree in Transportation Engineering from Tongji University, China, in 2016, and was a postdoc at California PATH, University of California, Berkeley, from 2016 to 2018. She is now is a researcher and team leader at California PATH, UC Berkeley. Her current research is focused on Deep Learning algorithms and applications for Autonomous Driving, including driving behavior learning, trajectory planning, control policy learning, and pedestrian intention prediction. She also collaborates with industries on projects such as intelligent traffic control and advanced vehicular technology assessment.
\end{IEEEbiography}
\vspace{-35pt}
\begin{IEEEbiography}[{\includegraphics[width=1in,height=1.25in,clip,keepaspectratio]{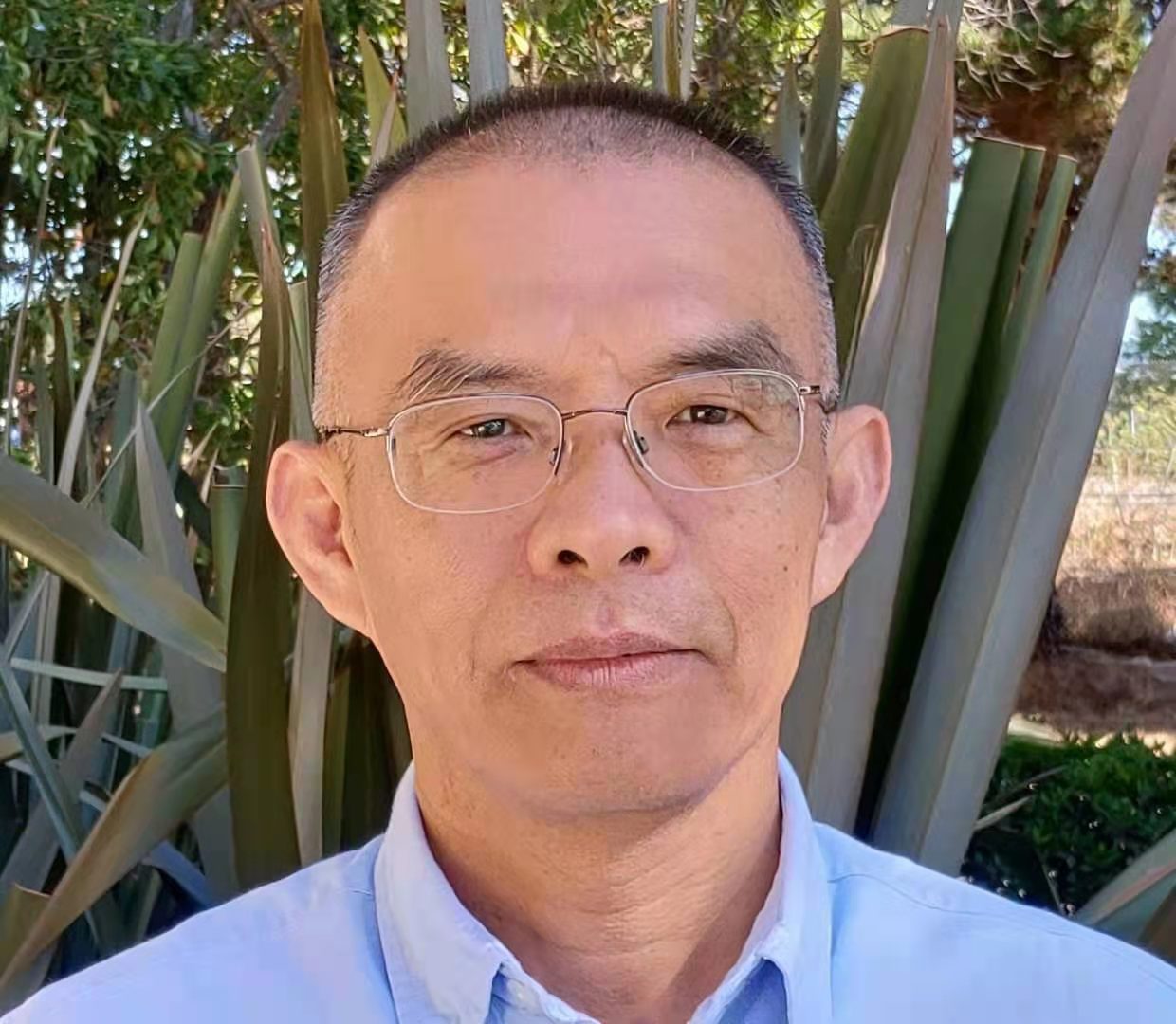}}]{Ching-yao Chan}
received his Ph.D. degree in Mechanical Engineering from University of California, Berkeley, US, in 1988. He has three decades of research experience in a broad range of automotive and transportation systems, and now is the Program Leader at California PATH and the Co-Director of Berkeley DeepDrive Consortium (bdd.berkeley.edu).He is now leading research in several topics, including driving behavior learning, pedestrian-vehicle interaction, sensor fusion for driving policy adaptation, and supervisory control in automated driving systems, human factors in automated driving.
\end{IEEEbiography}
\begin{IEEEbiography}[{\includegraphics[width=1in,height=1.25in,clip,keepaspectratio]{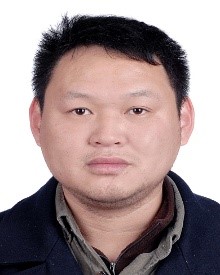}}]{Xiang Song}
received the Ph.D. degree from the School of Instrumentation Science and Engineering, Southeast University, China, in 2014. From 2014 to 2016, he was a Post-Doctoral Researcher with Southeast University. He is currently with the School of electronic engineering, Nanjing Xiaozhuang University of China. His research interests include sensor integration navigation, information fusion, and Intelligent Transportation System.
\end{IEEEbiography}
\vspace{-35pt}
\begin{IEEEbiography}[{\includegraphics[width=1in,height=1.25in,clip,keepaspectratio]{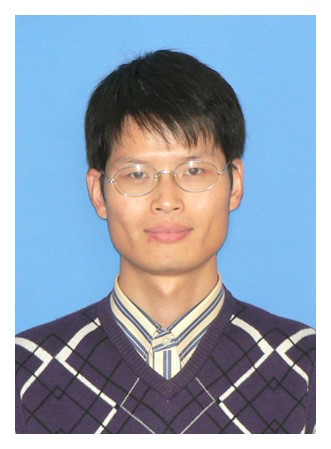}}]{Yang Chen}
received his Ph.D. degree in Harbin Engineering University, Harbi, China. He is now a associate professor at the School of Information Science and Engineering, Changzhou University. His research interests include underwater array signal processing and acoustic signal processing.
\end{IEEEbiography}






\end{document}